\documentclass[runningheads]{llncs}

\usepackage{eccv}

\usepackage{eccvabbrv}

\usepackage{graphicx}
\usepackage{booktabs}

\usepackage[accsupp]{axessibility}  

\usepackage{cite}
\usepackage{mathrsfs}
\usepackage{amsfonts}
\usepackage{amsmath}
\usepackage{amssymb}
\usepackage{bbding} 
\usepackage{booktabs}
\usepackage{tabularx}
\usepackage{multirow}
\usepackage{xcolor}
\usepackage[abs]{overpic}

\usepackage{algorithm}  
\usepackage{algorithmicx}  
\usepackage{algpseudocode}

\newcommand{\tabincell}[2]{\begin{tabular}{@{}#1@{}}#2\end{tabular}}

\def\eg{\emph{e.g}\onedot} 
\def\ie{\emph{i.e}\onedot} 
\def\etal{\emph{et al}\onedot}

\usepackage[pagebackref,breaklinks,colorlinks]{hyperref}

\usepackage{orcidlink}

\begin{document}

\title{Self-Supervised Video Desmoking for Laparoscopic Surgery}   

\titlerunning{SelfSVD}

\author{Renlong Wu\inst{1} \and
Zhilu Zhang\inst{1}$^{(}$\Envelope$^)$ 
\and
Shuohao Zhang\inst{1}
\and
Longfei Gou\inst{2}
\and
\\
Haobin Chen\inst{2}
\and    
Lei Zhang \inst{3}
\and
Hao Chen \inst{2}$^{(}$\Envelope$^)$ 
\and
Wangmeng Zuo \inst{1}
}

\authorrunning{R.Wu et al.}

\institute{
Harbin Institute of Technology, China \and
Southern Medical University, China \and
Hong Kong Polytechnic University, China
\\
\email{hirenlongwu@gmail.com,cszlzhang@outlook.com,
\\yhyzshrby@163.com,Calvin\_smu@163.com,Hao\-Bin\_Chen@outlook.com,
\\cslzhang@comp.polyu.edu.hk,chenhao.05@163.com,wmzuo@hit.edu.cn}
}

\maketitle

\begin{abstract}
Due to the difficulty of collecting real paired data, most existing desmoking methods train the models by synthesizing smoke, generalizing poorly to real surgical scenarios. Although a few works have explored single-image real-world desmoking in unpaired learning manners, they still encounter challenges in handling dense smoke. In this work, we address these issues together by introducing the self-supervised surgery video desmoking (SelfSVD). On the one hand, we observe that the frame captured before the activation of high-energy devices is generally clear (named pre-smoke frame, PS frame), thus it can serve as supervision for other smoky frames, making real-world self-supervised video desmoking practically feasible. On the other hand, in order to enhance the desmoking performance, we further feed the valuable information from PS frame into models, where a masking strategy and a regularization term are presented to avoid trivial solutions. In addition, we construct a real surgery video dataset for desmoking, which covers a variety of smoky scenes. Extensive experiments on the dataset show that our SelfSVD can remove smoke more effectively and efficiently while recovering more photo-realistic details than the state-of-the-art methods. The dataset, codes, and pre-trained models are available at \url{https://github.com/ZcsrenlongZ/SelfSVD}.

\end{abstract}

\section{Introduction}
\label{sec:intro}
Laparoscopy is employed to capture videos of the surgical sites to aid surgeons' decision-making, and it has found extensive application in the medical field~\cite{loukas2018video}.
However, during the surgery, the activation of high-energy devices (\eg, electrocautery and ultrasonic scalpel) leads to the destruction and vaporization of proteins and fats, as well as the evaporation of liquid water, inevitably causing smoke~\cite{Synchronizing}.
The smoke may obscure specific tissue details, consequently diminishing the quality of laparoscopic imaging and impeding surgeons in making informed judgments~\cite{azam2022smoke}.
Since it cannot be easily and quickly removed in vivo, post-processing laparoscopic images for desmoking~\cite{gu2015virtual} has become an effective and convenient manner to assist surgeons in observing the surgical sites clearly.
\begin{figure}[t]
\centering
\begin{overpic}[width=0.8\textwidth,grid=False]
{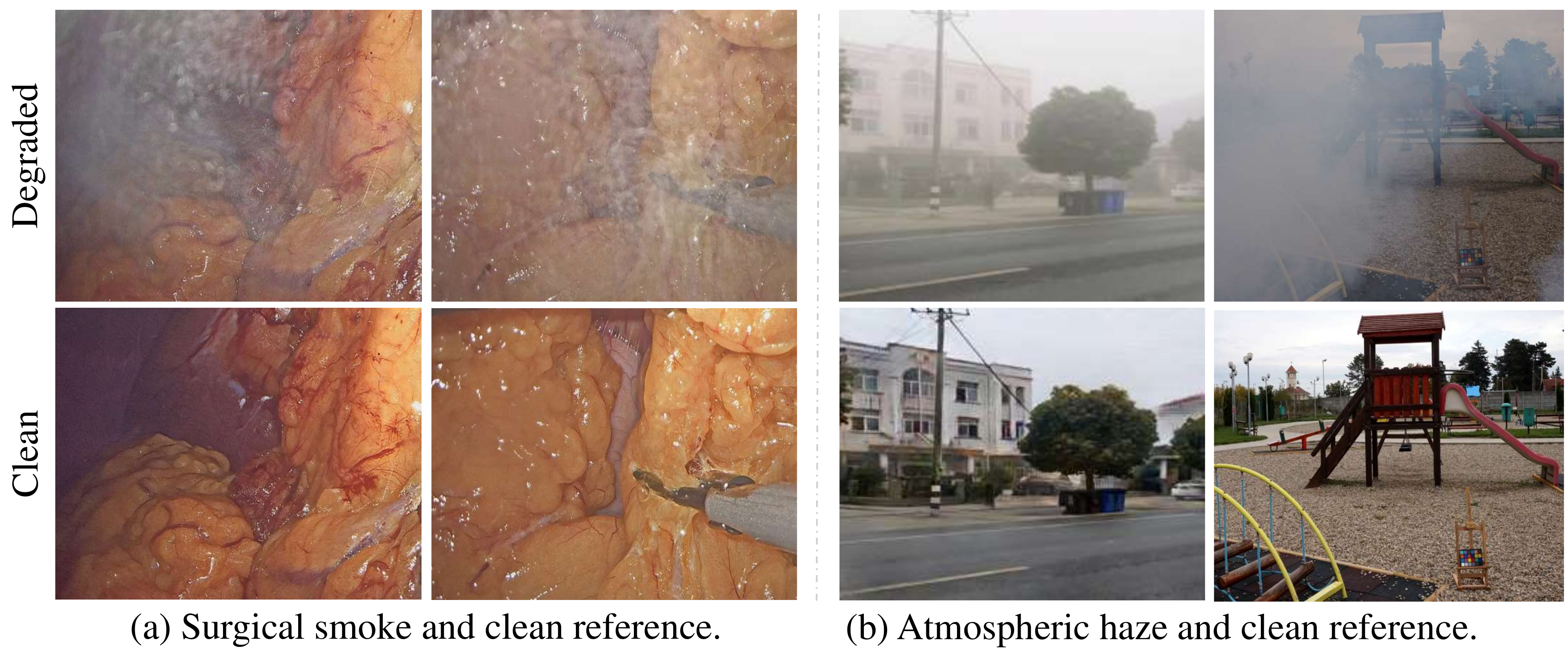}
\end{overpic}
\caption{The comparison of surgical smoke in (a) and atmospheric haze in (b). Degraded images are in the top row and their clean reference images are in the bottom row.}
\label{fig:intro_comparison}
\end{figure}

It is worth noting that although surgical smoke and atmospheric haze are somewhat similar in conformation, they are not exactly the same.
As shown in \cref{fig:intro_comparison}, the latter is usually locally homogeneous and follows the atmospheric scattering model, while the former is difficult to physically simulate and contains more diverse situations, such as mist, droplets, and streaks.
Consequently, the pre-trained dehazing models~\cite{fan2023non,cai2016dehazenet,li2017aod,guo2022image,chen2021psd} are generally ineffective for desmoking.
It is significant to give deliberate attention to both data and methods tailored for surgical smoke removal.

Recently, several attempts~\cite{wang2019multiscale,ma2021smoke,lin2021desmoking, zheng2023development,wang2023surgical} have been made towards desmoking, while still facing some challenges. 
In this task, collecting paired data is difficult and even infeasible.
To circumvent this problem, most existing methods employ a 3D graphics rendering engine~\cite{chen2019smokegcn, holl2020phiflow} to generate simulated smoky images~\cite{chen2019smokegcn,wang2023surgical,salazar2020desmoking} and videos~\cite{sengar2021multi}. 
However, the models trained on the synthetic data do not generalize well in real surgical scenarios, due to the domain gap between the synthetic smoke and the real-world one.
In addition, 
some works~\cite{venkatesh2020unsupervised,MARS-GAN,pan2022desmoke,salazarcolores2020desmoking,su2023multi} adopt unpaired learning manners~\cite{zhu2017unpaired,goodfellow2014generative} for real-world single-image smoke removal.
But they are unsatisfactory in handling dense smoke due to the inherent ill-posed nature.
Additionally, few real-world video desmoking methods are explored.

This work aims to bring surgery video desmoking into the real world, which is more practical significance.
Actually, sufficiently leveraging surgery video has the potential to address both the lack of real-world paired data and limited performance problems. 
On the one hand, the frames captured before the activation of high-energy devices usually have less smoke and similar contents as subsequent smoky ones.
We designate the almost clear frame as a pre-smoke (PS) frame. 
PS frame can provide effective supervision information for smoky frames. 
Thus, its reasonable utilization will make self-supervised real-world video desmoking feasible.
On the other hand, the surgical smoke level fluctuates over time, either intensifying or diminishing.
The neighboring frames may be complementary to the current one in smoke removal.
Thus, video desmoking is expected to outperform single-image desmoking significantly in alleviating ill-posed problems.

Specifically, we propose a novel self-supervised surgery video desmoking (SelfSVD) method in this work.
On the one hand, we utilize PS frame as misaligned supervision for smoky ones.
In order to calculate the loss function between desmoking output and supervision accurately, a pre-trained optical flow estimation network (\eg, PWC-Net~\cite{sun2018pwc}) is deployed to align the output with PS frame.
On the other hand, PS frame is also available during the test stage, and is particularly important for handling complex and dense smoke in the real world.
Thus, we further feed PS frame into models to help recover more details.
Unfortunately, such a manner can easily lead to trivial solutions.
To address this issue, we introduce a masking strategy and a regularization term. 
Besides, we suggest enhancing PS frame as better supervision by the above pre-trained SelfSVD model, and then taking it to fine-tune the model for improving visual effects.

We note that there is a dearth of real-world surgery video datasets for desmoking. 
In order to fill this gap, we collect multiple laparoscopic videos from professional hospitals and construct a laparoscopic surgery video desmoking (LSVD) dataset, which holds promise to benefit future studies.
Extensive experiments are conducted on the dataset.
The results show that our SelfSVD achieves better results than state-of-the-art methods in smoke removal and fine-scale detail recovery.
Furthermore, we also design a lightweight model that can be inferred in real-time.

The contributions can be summarized as follows:
\begin{itemize}
    \item We suggest utilizing the internal characteristics of real-world surgery videos for effective self-supervised video desmoking, and propose a SelfSVD solution, in which the pre-smoke (PS) frame  serves as an unaligned supervision. 
    
    \item We propose to take PS frame as an additional input to further improve desmoking performance, where a masking strategy and a regularization term are introduced to prevent trivial solutions.
    
    \item We construct a real-world laparoscopic surgery video desmoking (LSVD) dataset. Extensive experiments on the dataset demonstrate that our SelfSVD outperforms the state-of-the-art methods.
\end{itemize}

\section{Related Work}
\subsection{Supervised Desmoking}
Several studies~\cite{wang2018variational,zhu2015fast,tchaka2017chromaticity,wang2019multiscale,ma2021smoke, zheng2023development,wang2023surgical} have endeavored to address the surgical smoke removal problem.
Traditional approaches~\cite{zhu2015fast,tchaka2017chromaticity, wang2018variational} estimate the transmission components based on the atmospheric scattering model, but easily lead to color and structure artifacts. 
Later works~\cite{wang2019multiscale,ma2021smoke,lin2021desmoking, zheng2023development,wang2023surgical} adopt learn-based manners and design various networks based on convolutional neural networks (CNN) and vision transformers~\cite{dosovitskiy2020image}. 
For example, Lin \etal~\cite{lin2021desmoking} and Ma \etal~\cite{ma2021smoke} modify U-net~\cite{ronneberger2015u} architecture for surgical desmoking.
Wang \etal~\cite{wang2019multiscale} propose an encoder-decoder architecture with a Laplacian image pyramid decomposition strategy.
Wang \etal~\cite{wang2023surgical} employ Swin Transformer blocks~\cite{liang2021swinir} to enhance feature extraction.
Zheng \etal~\cite{zheng2023development} further develop a system jointly detecting and removing surgical smoke.
However, these methods generally simulate surgery smoke for training.
Due to the domain gap between the synthetic smoke and the real-world one, they can't generalize well to real surgery scenes.

\subsection{Supervised Dehazing}
Dehazing~\cite{cai2016dehazenet,li2017aod,guo2022image, dong2020multi, xiao2022single, li2018end,ren2018deep,zhang2021learning,liu2022phase,xu2023video,zheng2023curricular,qiu2023mb,guo2023dadfnet} is of great relevance to surgical desmoking, which aims to recover clean components from hazy ones affected by adverse weather conditions.
Li \etal ~\cite{li2017aod} reformulate the atmospheric scattering model and propose AODNet for image dehazing.
UHD~\cite{xiao2022single} introduces the infinite approximation of Taylor's theorem with Laplacian pyramid pattern.
DeHamer~\cite{guo2022image} further combine CNN~\cite{ronneberger2015u} and vison Transformer~\cite{dosovitskiy2020image} together for performance improvement.
Compared to the above single-image methods, video ones~\cite{li2018end,ren2018deep,zhang2021learning,zhang2021learning,liu2022phase,xu2023video} can leverage temporal clues between consecutive frames for more effective restoration.
Li \etal~\cite{li2018end} first build an united video detection and dehazing framework that focuses on temporal information fusion.
Ren \etal~\cite{ren2018deep} incorporate global semantic priors for smooth transmission map estimation.
Recently, MAPNet~\cite{xu2023video} explores physical haze priors to guide spatial information extraction and proposes a spatial-temporal alignment strategy to guide temporal features aggregation, achieving state-of-the-art performance.
However, employing their pre-trained models directly does not produce satisfactory desmoking results due to the discrepancy between atmospheric haze and surgical smoke.
Moreover, the lack of real-world paired data severely restricts the possibility of starting training from scratch.
In this work, we propose a self-supervised framework for video desmoking (SelfSVD) to address these issues.

\subsection{Unsupervised  Desmoking and Dehazing}
In contrast to supervised methods, unsupervised ones~\cite{Synchronizing,venkatesh2020unsupervised,MARS-GAN,pan2022desmoke,salazarcolores2020desmoking,chen2019smokegcn,su2023multi,li2021you,li2020zero,9366772,yang2018towards,li2022usid,engin2018cycle,yang2022self,fan2023non,chen2021psd} can be trained without paired supervision.
For surgical smoke removal, Cyclic-DesmokeGAN~\cite{venkatesh2020unsupervised} proposes a real-world image desmoking model based on CycleGAN~\cite{zhu2017unpaired}.
Desmoke-LAP~\cite{pan2022desmoke} and DCP-Pixel2Pixel~\cite{salazarcolores2020desmoking} introduce dark channel prior~\cite{he2010single} into loss and network design, respectively.
MS-CycleGAN~\cite{su2023multi} adapts a model pre-trained on synthetic data to real-world ones.

Apart from the above unsupervised desmoking methods, unsupervised dehazing ones~\cite{li2021you,li2020zero,9366772,yang2018towards,li2022usid,engin2018cycle,yang2022self} are also widely explored.
YOLY~\cite{li2021you} and ZID~\cite{li2020zero} perform dehazing in a zero-shot manner by disentangling the hazy image into the clean component and other ones.
RefinetNet~\cite{9366772}, DistentGAN~\cite{yang2018towards}, and USID-Net~\cite{li2022usid} build their models under GAN~\cite{goodfellow2014generative} framework.
$D^4$~\cite{yang2022self} converts the transmission map estimation into density and depth image prediction, achieving better results.
CycleDehaze~\cite{engin2018cycle} introduces a Laplacian pyramid network to handle high-resolution images efficiently.
However, these methods suffer from unstable training and fail to process dense smoke.
In contrast, our SelfSVD enables more stable and finer-scale restoration by sufficiently utilizing pre-smoke frames.
\begin{figure}[t!]
    \centering
    \includegraphics[width=0.7\linewidth]{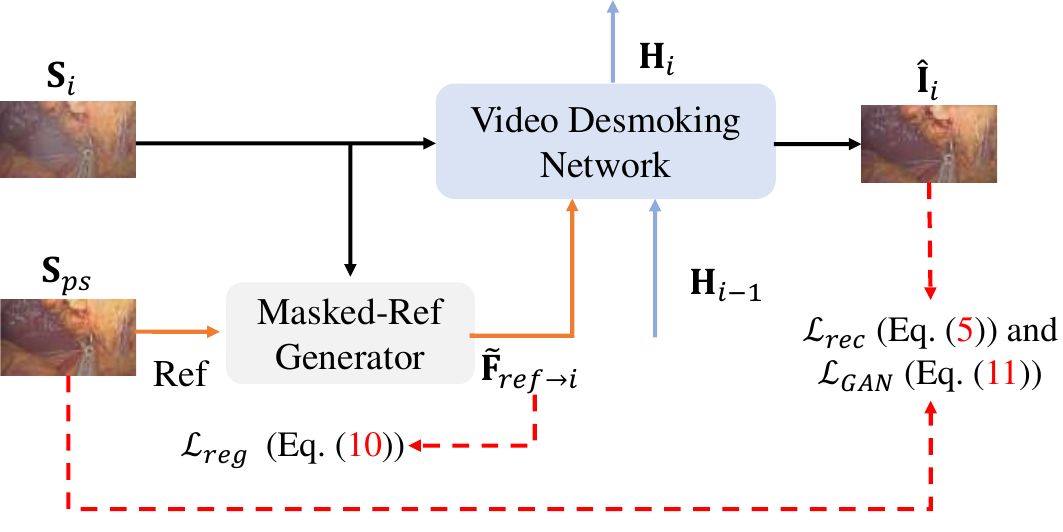}
    \caption{
    The illustration of processing the $i$-th smoky frame $\mathbf{S}_{i}$.
    PS frame ($\mathbf{S}_{ps}$) is taken as both supervision and reference (Ref) input.
    A masking strategy with the masked-ref generator as shown in \cref{fig:pipeline_2} and a regularization term as \cref{eq:RegLoss} are introduced to prevent trivial solutions.
    $\mathbf{H}_{i-1}$ is the temporal features from previous frames and $\mathbf{H}_{i}$ is the temporal features for subsequent ones.
    }
    \label{fig:pipeline_1}
\end{figure}

\section{Proposed Method}
We first describe the approach towards self-supervised learning, \ie, taking pre-smoke (PS) frame as supervision in Sec.~\ref{section:3_1}. 
Then we detail how to take PS frame as reference input and introduce the solutions to prevent trivial solutions in Sec.~\ref{section:3_2}.
Next, we introduce the way to enhance PS frame as better supervision for improving visual effects in Sec.~\ref{section:3_3}.
Finally, the details about the network architecture and learning objective are provided in Sec.~\ref{section:3_4}.

\subsection{Taking PS Frame as Supervision}
\label{section:3_1}
Given a smoky video consisting of $N$ frames $\{\mathbf{S}_i\}^{N}_{i=1}$, video desmoking aims to restore the corresponding clean components ${\{\mathbf{\hat{I}}_i\}}^{N}_{i=1}$ with temporal clues, \ie,
\begin{equation}
\label{eq:baseline}
\{{\mathbf{\hat{I}}_i\}}^{N}_{i=1} = \mathcal{D}({\{\mathbf{S}_i\}^{N}_{i=1}}; \Theta_{\mathcal{D}}),
\end{equation}
where $\mathcal{D}$ denotes the video desmoking model with the parameter $\Theta_\mathcal{D}$.
However, the paired smoky-clean videos are difficult to acquire in the real world. 
Several studies~\cite{wang2018variational,zhu2015fast,tchaka2017chromaticity,wang2019multiscale,ma2021smoke, zheng2023development,wang2023surgical} utilize synthetic data for training, but the domain gap between synthetic smoke and real one hinders their effective applications in the real world.
A few single-image unpaired methods~\cite{venkatesh2020unsupervised,MARS-GAN,pan2022desmoke,salazarcolores2020desmoking,su2023multi} are explored to overcome this issue, but their training is less stable and their ill-posed nature makes them less effective in processing dense smoke.
In contrast to these approaches, we propose to utilize the internal characteristics of real-surgery video for self-supervised video desmoking to address these issues together, as shown in \cref{fig:pipeline_1}.
Specifically, we note that the pre-smoke frame (PS frame, $\mathbf{S}_{ps}$) is clearer than subsequent smoky ones and can be regarded as their supervision.
\begin{figure}[t!]
    \centering
    \begin{overpic}[width=0.9\textwidth,grid=False]
    {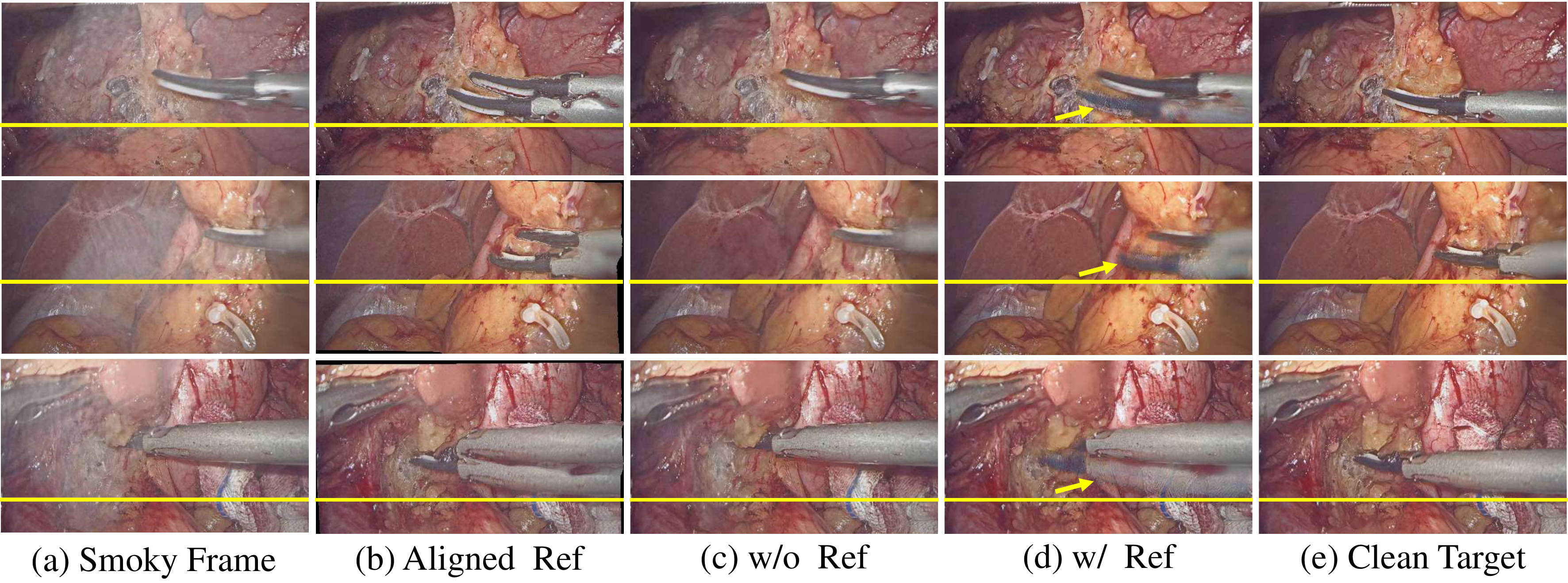}
    \end{overpic}
    \caption{Examples of trivial solutions. When inputting PS frame as Ref naively, the imperfect optical flow between Ref and the smoky frame leads to trivial solutions, as indicated by yellow arrows.
    The same positions are marked with yellow lines.}
    \label{fig:overfiting}
\end{figure}

Motivated by this, we introduce SelfSVD to achieve self-supervised learning, where we can train the video desmoking network $\mathcal{D}$ as,
\begin{equation}
\label{eq:selfsvd}
\Theta_{\mathcal{D}}^* = 
\arg \min _{\Theta_{\mathcal{D}}} \mathcal{L}
\left(\mathcal{D}
(\{\mathbf{S}_i\}^{N}_{i=1}; 
\Theta_{\mathcal{D}}), 
\mathbf{S}_{ps}\right),
\end{equation}
where $\mathcal{L}$ denotes the learning objective.
It is worth noting that ${\mathbf{\hat{I}}_i}$ is not spatially aligned with $\mathbf{S}_{ps}$.
Taking $\mathbf{S}_{ps}$ as supervision directly may lead to blurry results~\cite{zhang2021learning,SelfDZSR,li2023learning}.
Instead, we adopt a deformation-based learning objective to tolerate the misalignment.
In particular, a pre-trained optical flow network $\mathcal{O}$ is first used to estimate the optical flow $\mathbf{\Psi}_{ps \rightarrow i}$ from $\mathbf{S}_{ps}$ to $\mathbf{\hat{I}}_{i}$, \ie,
\begin{equation}
\mathbf{\Psi}_{ps \rightarrow i} = \mathcal{O}(\mathbf{S}_{ps}, \mathbf{\hat{I}}_{i}).
\end{equation}
Then $\mathbf{\hat{I}}_{i}$ is back warped towards $\mathbf{S}_{ps}$ with a warping operation $\mathcal{W}$ according to the estimated optical flow $\mathbf{\Psi}_{ps \rightarrow i}$, \ie,
\begin{equation}
\mathbf{\hat{I}}_{i \rightarrow ps} = \mathcal{W}(\mathbf{\hat{I}}_i, \mathbf{\Psi}_{ps \rightarrow i}).
\end{equation}
Finally, $\mathbf{\hat{I}}_{i \rightarrow ps}$ is spatially aligned with $\mathbf{S}_{ps}$. The reconstruction loss $\mathcal{L}_{rec}$ of the desmoking model can be written as,
\begin{equation}
\label{eq:warpL1}
\mathcal{L}_{rec} = \sum_{i=1}^N||\mathbf{V}_{i} \odot ( \mathbf{\hat{I}}_{i \rightarrow ps} - \mathbf{S}_{ps})||_1.
\end{equation}
$\odot$ is a pixel-wise multiply operation.
$\mathbf{V}_{i}$ is a mask that indicates the valid positions of optical flow.
The $j$-th element of $\mathbf{V}_{i}$ can be calculated as,
\begin{equation}
\mathbf{V}_{i}^{j} = \mathbf{\operatorname{sgn}} 
\left( \max \left(0, 
[\mathcal{W}(\mathbf{1}, \mathbf{\Psi}_{ps \rightarrow i})]_j - \tau \right)\right).
\end{equation}
$\mathcal{W}$ is warping operation.
$\max$ and $\operatorname{sgn}$ are the maximum and sign function, respectively.
$\mathbf{1}$ is an all-1 matrix, 
$[\cdot]_j$ denotes the $j$-th element.
$\tau$ is set to 0.999.

\begin{figure}[t!]
    \centering
    \includegraphics[width=0.9\linewidth]{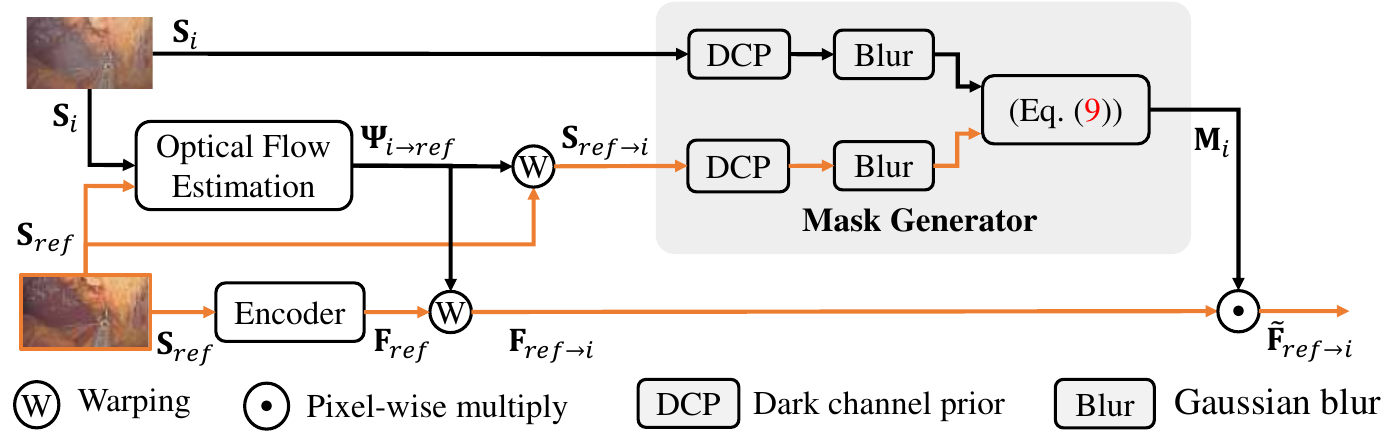}
    \caption{
    The structure of masked-ref generator.
    The mask generator is used to generate a mask $\mathbf{M}_i$, which is employed to produce the masked reference features $\tilde{\mathbf{F}}_{ref \rightarrow i}$.
    }
    \label{fig:pipeline_2}
\end{figure}
\subsection{Taking PS Frame as Reference Input}
\label{section:3_2}
We note that PS frame is available during both training and testing. 
Thus, beyond taking PS frame as supervision, we can further take the PS frame as a reference (Ref) input to guide smoke removal. 
Denote by $S_{ref}$ the Ref (\ie, $S_{ps}$), \cref{eq:selfsvd} can be modified as,
\begin{equation}
\Theta_{\mathcal{D}}^* = 
\arg \min _{\Theta_{\mathcal{D}}} \mathcal{L}
\left(\mathcal{D}
(\{\mathbf{S}_i\}^{N}_{i=1},  \mathbf{S}_{ref}; 
\Theta_{\mathcal{D}}), 
\mathbf{S}_{ps}\right).
\end{equation}
Denote the features of Ref and the $i$-th smoky frame by $\mathbf{F}_{ref}$ and $\mathbf{F}_{i}$ respectively (see \cref{section:3_4} for details).
The spatial misalignment between them should be addressed first.
Thus, we estimate the optical flow $\mathbf{\Psi}_{i \rightarrow ref}$ from the $i$-th smoky frame $\mathbf{S}_{i}$ to $\mathbf{S}_{ref}$.
Then, we back warp $\mathbf{F}_{ref}$ to $\mathbf{F}_{i}$ according to  $\mathbf{\Psi}_{i \rightarrow ref}$, generating the warped reference features $\mathbf{F}_{ref\rightarrow i}$.
When $\mathbf{\Psi}_{i \rightarrow ref}$ is perfectly estimated, $\mathbf{F}_{ref\rightarrow i}$ is naturally perfectly aligned with $\mathbf{F}_{i}$.
In this case, the desmoking model can produce better results.
However, it is not realistic to achieve perfect optical flow estimation due to the interference of surgical smoke and content occlusion.
Some contents (\eg, areas where high-energy devices move significantly) in $\mathbf{F}_{ref}$ may not be correctly mapped to the corresponding positions in $\mathbf{F}_{i}$, being kept in $\mathbf{F}_{ref\rightarrow i}$.
Moreover, the desmoking model tends to utilize features from $\mathbf{F}_{ref\rightarrow i}$ rather than $\mathbf{F}_{i}$, as the former contains more clean information relevant to supervision.
Thus, the model easily over-fits the Ref, as shown in \cref{fig:overfiting}.
To circumvent the issue, we introduce a hard masking strategy and supplement a soft regularization term, as described below.

\noindent \textbf{Masking Strategy.} 
In regions with significantly inaccurate optical flow, we prefer to forget the reference information to avoid output contents being inconsistent with smoky inputs.
As shown in \cref{fig:pipeline_2}, we suggest generating a mask $\mathbf{M}_{i}$ to indicate these regions in $\mathbf{F}_{ref\rightarrow i}$, and exclude corresponding features, \ie,
\begin{equation}
\label{eq:mask_ref_qen}
\tilde{\mathbf{F}}_{ref\rightarrow i} = \mathbf{M}_{i} \odot \mathbf{F}_{ref\rightarrow i}.
\end{equation}
As shown in \cref{fig:pipeline_2}, to obtain $\mathbf{M}_{i}$, we first align $\mathbf{S}_{ref}$ to $\mathbf{S}_{i}$, obtaining the warped reference image $\mathbf{S}_{ref \rightarrow i}$.
Then, we process $\mathbf{S}_{ref \rightarrow i}$ and $\mathbf{S}_{i}$ with DCP~\cite{he2010single} and large-kernel Gaussian blur operations to alleviate the disturbance of surgery smoke.
Next, we divide $\mathbf{S}_{ref \rightarrow i}$ and $\mathbf{S}_{i}$ into $P$ non-overlapping patches respectively, \ie, $\{{\mathbf{S}_{ref \rightarrow i}^{p}\}}^{P}_{p=1}$ and $\{{\mathbf{S}_{i}^{p}\}}^{P}_{p=1}$.
For each pair $\mathbf{S}_{ref \rightarrow i}^{p}$ and $\mathbf{S}_{i}^{p}$, their structure should be significantly distinct when optical flow is estimated inaccurately.
So we adopt structural similarity (SSIM) metric~\cite{wang2004image} to detect such patch, \ie,
\begin{equation}
m_{i}^{p} = \mathbf{\operatorname{sgn}}
\left(\max \left(0,
        \texttt{SSIM} ( \mathbf{S}_{ref \rightarrow i}^{p}, \mathbf{S}_{i}^{p} )
        -\mathbf{\epsilon}\right)\right).
\end{equation}
$m_{i}^{p}$ is the value of $\mathbf{M}_{i}$ in the $p$-th patch.
$\epsilon$ is a threshold and is set to 0.92.
The patch size is set to $8$.
Please see more details about mask in the Suppl.

\begin{figure}[t!]
    \centering
    \includegraphics[width=0.76\linewidth]{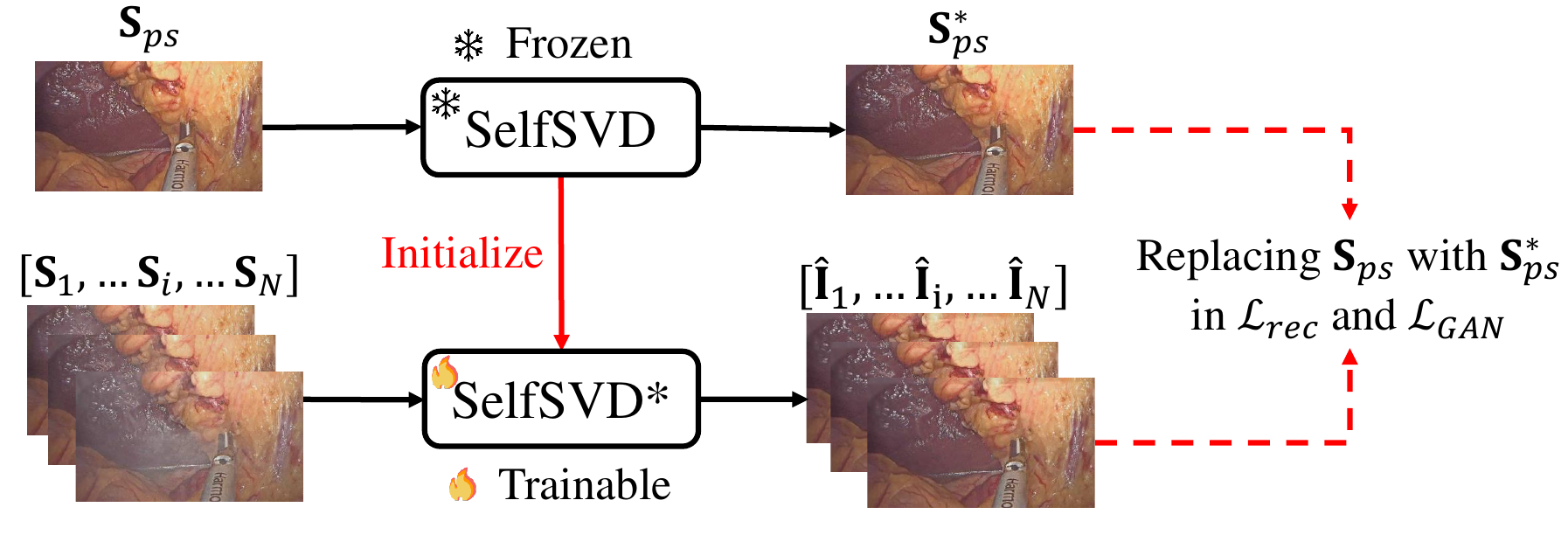}
    \vspace{-2mm}
    \caption{
    The illustration of enhancing PS frame ($\mathbf{S}_{ps}$) as supervision.
    We regard $\mathbf{S}_{ps}$ as a frame with less smoke and feed it into a pre-trained SelfSVD model, generating a cleaner result $\mathbf{S}_{ps}^{\ast}$.
    Then, $\mathbf{S}_{ps}^{\ast}$ is taken as improved supervision to fine-tune the SelfSVD model by $\mathcal{L}_{rec}$ and $\mathcal{L}_{GAN}$  (\ie, replacing $\mathbf{S}_{ps}$ with $\mathbf{S}_{ps}^{\ast}$ in \cref{eq:warpL1} and \cref{eq:GANLoss}), getting an improved model named SelfSVD$^{\ast}$.
    }
    \label{fig:self_svd_star}
\end{figure}

\noindent \textbf{Regularization Strategy.}
The areas where optical flow is slightly wrong may lead to potential trivial solutions, as they are not easily detected explicitly.
We use $\ell_1$ regularization~\cite{donoho2006compressed,bishop2006pattern} to constraint  $\mathbf{F}_{ref \rightarrow i}$ in these areas (indicated by mask $\mathbf{M}_{i}$) to be sparse.
The regularization loss $\mathcal{L}_{reg}$ can be written as,
\begin{equation}
\label{eq:RegLoss}
\mathcal{L}_{reg} = || \mathbf{M}_{i} \odot \mathbf{F}_{ref \rightarrow i} ||_{1}.
\end{equation}

\subsection{Enhancing PS Frame as Supervision}
\label{section:3_3}
For model training, there are two possible ways to introduce PS frame.
One is to always adopt the starting frame in surgery, and another one is to select it dynamically as the surgery proceeds. 
We adopt the latter way, as the video contents may change significantly, causing starting frames to provide insufficient information for long-distance ones.
However, in this manner, some smoke generated by previous activation of high-energy devices may remain in $\mathbf{S}_{ps}$.
Thus, naively taking $\mathbf{S}_{ps}$ as supervision may not be the optimal solution.

Taking this into account, we improve the training of SelfSVD by enhancing $\mathbf{S}_{ps}$ as better supervision, dubbed SelfSVD$^\ast$, as shown in \cref{fig:self_svd_star}. 
In particular, a SelfSVD model is first pre-trained under the supervision of $\mathbf{S}_{ps}$.
Then, we regard $\mathbf{S}_{ps}$ as a frame with less smoke and feed it into the pre-trained SelfSVD, getting a cleaner result $\mathbf{S}_{ps}^{\ast}$.
Finally, a SelfSVD$^\ast$ model is obtained by fine-tuning the pre-trained SelfSVD with $\mathbf{S}_{ps}^{\ast}$ as supervision, and the original SelfSVD model can be detached during testing.
%

\subsection{Network Architecture and Learning Objective}
\label{section:3_4}
\noindent \textbf{Network Architecture.}
Considering the practical application, the desmoking model needs to process surgery videos online, rather than offline. 
Therefore, we design a video desmoking network based on the unidirectional recurrent neural network~\cite{wu2023rbsr}.
According to the function of each component, it can be divided into five modules, \ie, feature encoder, masked-ref generator, alignment, fusion, and reconstruction module.
When processing the $i$-th smoky frame $\mathbf{S}_{i}$,
we first pass $\mathbf{S}_{i}$ to the encoder to obtain its features $\mathbf{F}_{i}$.
$\mathbf{S}_{ref}$ and $\mathbf{S}_{i}$ are fed into the masked-ref generator to get the masked reference features $\tilde{\mathbf{F}}_{ref\rightarrow i}$, which has been introduced in \cref{section:3_2}.
Then, we deploy the alignment module based on the optical flow to align previous temporal features $\mathbf{H}_{i-1}$ to $\mathbf{F}_{i}$, getting the warped ones $\mathbf{H}_{i-1 \rightarrow i}$.
Finally, the fusion module takes $\mathbf{F}_{i}$, $\tilde{\mathbf{F}}_{ref\rightarrow i}$ and $\mathbf{H}_{i-1 \rightarrow i}$ as inputs to get the fused feature representations, which are passed to the reconstruction module for generating the restored clean component $\mathbf{\hat{I}}_{i}$.
Please see more details about the network architecture in the Suppl.
\noindent \textbf{Learning Objective.}
To further improve the visual quality, we adopt adversarial loss~\cite{mao2017least} to train our desmoking networks, which can be written as,
%
\begin{equation}
\label{eq:GANLoss}
\mathcal{L}_{GAN}=\frac{1}{2} \mathbb{E}_{\mathbf{S} 
\sim {\mathcal{P}_{\mathbf{S}}}} 
\left[
\mathcal{DISC}
(\mathcal{D}(\mathbf{S}, \mathbf{S}_{ref}))-1
\right]^2,
\end{equation}
where $\mathbf{S}$ denotes the smoky video $\{\mathbf{S}_i\}^{N}_{i=1}$, and $\mathcal{DISC}$ is the discriminator~\cite{zhu2017unpaired} (see the Suppl. for detailed structure).
The discriminator is trained by,
\begin{equation}
    \begin{split}
    \mathcal{L}_{DISC}=\frac{1}{2} \mathbb{E}_{\mathbf{S}_{ps} \sim \mathcal{P}_{\mathbf{S}_{ps}}}
        [\mathcal{DISC}(\mathbf{S}_{ps})-1]^2
        \\
        +
        \frac{1}{2} 
        \mathbb{E}_{\mathbf{S} \sim \mathcal{P}_{\mathbf{S}}}
        [\mathcal{DISC}(\mathcal{D}(\mathbf{S}, \mathbf{S}_{ref}))]^2
    \end{split}.
\end{equation}
Overall, combined \cref{eq:warpL1}, \cref{eq:RegLoss} and \cref{eq:GANLoss}, the loss terms of SelfSVD can be written as,
\begin{equation}
\label{eq:allLoss}
\mathcal{L} = 
\mathcal{L}_{rec} + \lambda_{reg}\mathcal{L}_{reg} + \lambda_{GAN}\mathcal{L}_{GAN},
\end{equation}
where $\lambda_{reg}$ and $\lambda_{GAN}$ are set to 0.05 and 1.0, respectively.
When training SelfSVD$^*$ model, $\mathbf{S}_{ps}$ in \cref{eq:allLoss} is replaced with $\mathbf{S}_{ps}^*$, as stated in \cref{section:3_3}.

\section{LSVD Dataset}
\label{sec:dataset}
Cyclic-DesmokeGAN~\cite{venkatesh2020unsupervised} and Desmoke-LAP~\cite{pan2022desmoke} collect 1,200 smoky images and 3,000 ones from cholecystectomy and hysterectomy surgery recordings, respectively.
However, they are only used for single-image desmoking, being unsuitable for our video desmoking.
Sengar \etal~\cite{sengar2021multi} propose a video desmoking dataset but the smoke is simulated by the 3D graphics rendering engine~\cite{holl2020phiflow}.
As far as we know, real-world laparoscopic surgery video desmoking datasets are currently scarce, which  implies a high demand for this dataset.

In this work, we construct the LSVD dataset by collecting laparoscopic surgery videos of 40 patients from professional hospitals. 
We first select the frames manually where surgeons start activating the high-energy devices in each video.
Then, we take its preceding frame as PS frame and several subsequent ones (until the one in which smoke is nearly dismissed) as smoky frames.
Finally, 486 video clips are collected, where each clip contains 20$\sim$50 frames with a resolution of $1080 \times 1920$.
416 clips are used for the training set, and the remaining 70 ones are used for the testing set.
The dataset covers diverse and complex real-world surgical smoke.
We provide some examples in the Suppl.

\section{Experiments}

\subsection{Implementation Details}
For optical flow estimation, we adopt a pre-trained PWC-Net~\cite{sun2018pwc}.
During training, we randomly crop patches and augment them with random flips. 
The batch size is set to 4 and the patch size is set to $256 \times 256$.
SelfSVD is trained with ADAM optimizer ~\cite{kingma2014adam}
with ${\beta}_{1} = 0.9$ and ${\beta}_{2} = 0.99$ for 100k iterations.
Cosine annealing strategy ~\cite{loshchilov2016sgdr} is employed to decrease the learning rate from $1 \times 10^{-4}$ to $1 \times 10^{-7}$ steadily.
For the training of SelfSVD$^{\ast}$, we fine-tune the pre-trained SelfSVD model for additional 40k iterations and set the initial learning rate to $5 \times 10^{-5}$.
All experiments are conducted with PyTorch~\cite{paszke2019pytorch} on a single Nvidia GeForce RTX A6000 GPU.

SelfSVD and SelfSVD$^\ast$ keep the same number of residual blocks~\cite{he2016deep} as BasicVSR++~\cite{chan2022basicvsr++}, and the computation costs are generally similar to the video processing methods.
Moreover, to make the model inference cost consistent with some single-image processing methods, we present the lightweight versions by reducing the numbers and channels of residual blocks, dubbed SelfSVD-S and SelfSVD$^\ast$-S respectively.
Please see more details in the Suppl.

\begin{table*}[t!] 
    \small
    \renewcommand\arraystretch{1}
    \begin{center}
	\caption{Quantitative comparison on LSVD dataset.  $\uparrow$ denotes the higher metric the better, and $\downarrow$ denotes the lower one the better. The best results in each category are marked in \pmb{bold}.} 
	\label{tab:sota_comparison}
        \scalebox{0.8}{
	\begin{tabular}{clcccccccc}
		\toprule
		 & Methods & PSNR$\uparrow$ & SSIM$\uparrow$ & FADE$\downarrow$ & NIQE$\downarrow$ & PI$\downarrow$ & \tabincell{c}{\#FLOPs \\ (G)} & \tabincell{c}{\#Time \\ (s)} & \tabincell{c}{\#Params \\ (M)}\\
        \midrule
       \multirow{2}{*}{\begin{tabular}[c]{@{}c@{}} Unsupervised \\ Image Processing  \end{tabular}} 
        & PSD~\cite{chen2021psd} & 14.69 & 0.3722 & \pmb{0.1715} & \pmb{5.44} & 4.37 & 905 & 0.31 & 31.11\\
        \multicolumn{1}{c}{}
                & DCP~\cite{he2010single} & \pmb{17.77} & \pmb{0.5582} & 0.3217 & 5.53 & \pmb{4.25} & - & \pmb{0.08} & -\\
        \midrule
        \multirow{4}{*}{\begin{tabular}[c]{@{}c@{}} Unpaired \\ Image Processing  \end{tabular}}
                & DCP-Pixel2Pixel~\cite{salazarcolores2020desmoking} & 19.65 & 0.5079 & 0.5509 & 6.30 &4.50 & \pmb{195} & \pmb{0.02} & 54.40\\

        \multicolumn{1}{c}{} 
                & DistentGAN~\cite{shyam2021towards} & 21.51 & 0.6037 & 0.6135 & 4.95 & \textbf{3.60} & 1572 & 0.22 & 11.38\\
        \multicolumn{1}{c}{} 
                & Desmoke-LAP~\cite{pan2022desmoke} & 22.52 & 0.6170 & \pmb{0.5386} & \pmb{4.75} & 3.93 & 1571 & 0.15 & 11.38\\ 
        \multicolumn{1}{c}{} 
                & RefineNet~\cite{9366772} & \pmb{22.87} & \pmb{0.6177} & 0.5749 & 6.93 & 4.13 & 1780 & 0.34 & \textbf{0.85}\\ 
        \midrule
        \multirow{7}{*}{\begin{tabular}[c]{@{}c@{}} Self-Supervised \\Image Processing  \end{tabular}} 
        & UHD~\cite{xiao2022single} & 20.93 & 0.5792 & 1.3573 & 6.26 & 6.18 & 123 & 0.27 & 34.55\\
       \multicolumn{1}{c}{} 
        & MSDesmoking~\cite{wang2019multiscale} & 21.53 & 0.5997 & 1.1290 & \pmb{5.20} & 5.19 & 605 & 0.04 & 8.80\\
        \multicolumn{1}{c}{}
                & Wang \etal~\cite{wang2023surgical} & 22.65 & 0.6078 & 1.1202 & 7.63 & 7.11 & 6584 & 28.20 & 3.19\\  
        \multicolumn{1}{c}{} & MSBDN~\cite{dong2020multi} & 22.69 & 0.6093 & 0.8995 & 6.07 & 5.40 & 779 & 0.29 & 31.35\\
        \multicolumn{1}{c}{} 
        & AODNet~\cite{li2017aod} & 23.03 & 0.6116 & 1.0417 & 5.96 & 4.79 & \pmb{4} & 0.08 & \pmb{0.002}\\
        \multicolumn{1}{c}{} 
                & DADFNet~\cite{guo2023dadfnet} & 23.06 & 0.6143 & \pmb{0.4948} & 5.29 & \pmb{4.08} & 198 & \pmb{0.03} & 0.85\\      
       \multicolumn{1}{c}{} 
                & Dehamer~\cite{guo2022image} & \pmb{23.13} & \pmb{0.6184} & 1.1508 & 6.93 & 6.36 & 1564 & 0.41 & 132.45\\

        \midrule
        \multirow{7}{*}{\begin{tabular}[c]{@{}c@{}} Self-Supervised \\ Video Processing  \end{tabular}} 
                & BasicVSR~\cite{chan2021basicvsr} & 23.00 & 0.6168 & 0.5609 & 5.60 & 4.13  & 831 & 0.13 & 6.29\\
        \multicolumn{1}{c}{} 
                & BasicVSR++~\cite{chan2022basicvsr++} & 23.35 & 0.6196 & 0.5665 & 5.50 & 3.90 & 1197 & 0.19 & 9.76\\
        \multicolumn{1}{c}{} 
                & MAPNet~\cite{xu2023video} & 23.28 & 0.6152 & 0.9632 & 5.53 & 5.43 & 261 & 0.20 & 28.75\\
                 & (Ours) SelfSVD-S & 23.84 & 0.6183 & 0.4787 & 5.04 & 3.94 & \pmb{169} & \pmb{0.03} & \pmb{1.92}\\ 
                & (Ours) SelfSVD$^\ast$-S& 24.00 & 0.6209 & 0.4404 & 4.87 & 3.92 & \pmb{169}  & \pmb{0.03} & \pmb{1.92}\\      
                & (Ours) SelfSVD & 24.23 & 0.6216 & 0.4626 & 4.85 & 3.87 & 996  & 0.18 & 15.58\\
        \multicolumn{1}{c}{} 
                & (Ours) SelfSVD$^\ast$ & \pmb{24.58} & \pmb{0.6279} & \pmb{0.4193} & \pmb{4.72} & \pmb{3.86} & 996 & 0.18 & 15.58\\ 
		\bottomrule
	\end{tabular}}
    \end{center}
\end{table*}

\subsection{Evaluation Configurations}
On the one hand, as smoky frames have no paired clean frames, we have to utilize PS frame to evaluate desmoking methods. 
To remove the possible smoke in PS frame, we first feed it into a pre-trained SelfSVD model and take the pre-processed PS frame as the clean target. 
Then we adopt the aligned PSNR~\cite{huynh2008scope} and SSIM~\cite{wang2004image} as the reference evaluation metrics, following similar works~\cite{bhat2021deep,dudhane2022burst,wu2023rbsr}
Specifically, we align desmoking results to the target by a pre-trained optical flow network (\ie, PWC-Net~\cite{sun2018pwc}) and calculate PSNR and SSIM between aligned results and the target.
Simultaneously, we also report the metrics when taking the original PS frame as the target in the Suppl.
On the other hand, we employ no-reference metrics (\ie, FADE~\cite{choi2015referenceless}, NIQE~\cite{mittal2012making}, and PI~\cite{blau20182018}) to assess the desmoking results.
In addition, the number of model parameters and FLOPs, as well as inference time per frame when inputting $1080 \times 1920$ videos are also reported.

\subsection{Comparison with State-of-the-Art Methods}
%
%
Related learning-based methods are either trained on synthetic data in a supervised manner\cite{li2017aod,wang2019multiscale,dong2020multi,xiao2022single,guo2022image,guo2023dadfnet,chan2021basicvsr,chan2022basicvsr++,xu2023video}, or trained in a unpaired manner\cite{salazarcolores2020desmoking,shyam2021towards,9366772,pan2022desmoke}.
For fair comparisons, we retrain their models on our LSVD dataset.
For training supervised methods, there is no paired ground-truth.
Instead, we take the aligned PS frame as the supervision, which is the same as the setting of training our models as stated in Sec.~\ref{section:3_1}.
Thus, they become self-supervised methods. 
For unpaired learning methods, we take the original PS frame as their unpaired supervision.
Overall, we compare our methods (\ie, SelfSVD-S and SelfSVD$^\ast$-S, SelfSVD, SelfSVD$^\ast$) against 16 related state-of-the-art ones, including 2 unsupervised image processing methods (\ie, PSD~\cite{chen2021psd}, DCP~\cite{he2010single}),
4 unpaired image ones (\ie, DCP-Pixel2Pixel~\cite{salazarcolores2020desmoking},  DistentGAN~\cite{shyam2021towards}, and Desmoke-LAP~\cite{pan2022desmoke}, RefineNet~\cite{9366772}), 7 self-supervised image ones (\ie, 
UHD~\cite{xiao2022single},
MSDesmoking~\cite{wang2019multiscale},
Wang \etal~\cite{wang2023surgical}
MSBDN~\cite{dong2020multi},
AODNet~\cite{li2017aod},     DADFNet~\cite{guo2023dadfnet},
Dehamer~\cite{guo2022image},) and 3 self-supervised video ones (\ie, BasicVSR~\cite{chan2021basicvsr}, BasicVSR++~\cite{chan2022basicvsr++}, and MAPNet~\cite{xu2023video}).
We do not compare with unsupervised and unpaired video ones, as few works explore that as far as we know.
Beyond conducting experiments on our dataset, we provide more comparisons on a synthetic dataset in the Suppl. 

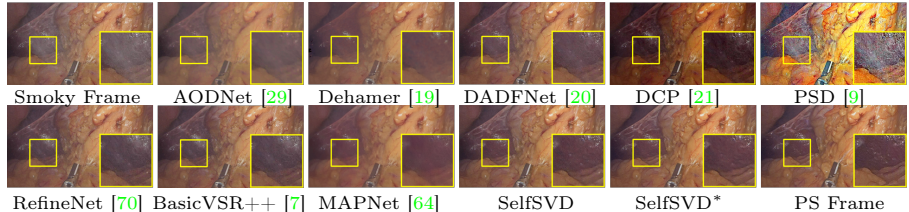
\begin{figure*}[t!]
    \centering
    \begin{subfigure}{0.99\textwidth}
        \ContinuedFloat
        \begin{overpic}[width=0.99\textwidth,grid=False]
        {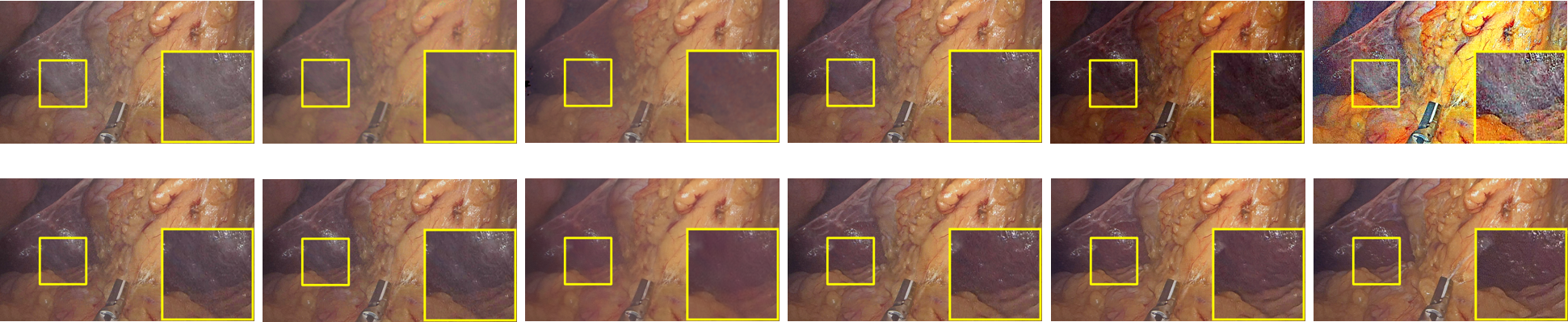}
          \put(0,32){\scriptsize{ Smoky Frame}}
          \put(60,32){\scriptsize{ AODNet~\cite{li2017aod}}}
          \put(115,32){\scriptsize{ Dehamer~\cite{guo2022image}}}
          \put(170,32){\scriptsize{ DADFNet~\cite{guo2023dadfnet}}}
          \put(235,32){\scriptsize{ DCP~\cite{he2010single}}}
          \put(295,32){\scriptsize{ PSD~\cite{chen2021psd}}}

          \put(2,-8){\scriptsize{RefineNet~\cite{9366772}}}
          \put(55,-8){\scriptsize{BasicVSR++~\cite{chan2022basicvsr++}}}
          \put(115,-8){\scriptsize{ MAPNet~\cite{xu2023video}}}
          \put(183,-8){\scriptsize{ \scriptsize{SelfSVD}}}
          \put(235,-8){\scriptsize{ \scriptsize{SelfSVD$^*$}}}
          \put(295,-8){\scriptsize{ \scriptsize{PS Frame}}}
        \end{overpic}
    \end{subfigure}

    
    \caption{Qualitative comparison on LSVD dataset. Our methods generate results more consistent with the PS frame. } 
    \label{fig:visualize_comparison_1}
    \vspace{-4mm}
\end{figure*}

    

\noindent\textbf{Quantitative Comparison.}
\cref{tab:sota_comparison} summarizes the quantitative results. 
First, video processing methods generally perform better than single-image ones, which demonstrates the benefits of utilizing temporal clues for surgical smoke removal.
Second, our methods outperform state-of-the-art ones by a large margin.
Among all competing methods, BasicVSR++~\cite{chan2022basicvsr++} has achieved the best PSNR score of 23.35dB.
Our SelfSVD and SelfSVD$^{\ast}$ achieve PSNR gains of 0.88dB and 1.23dB over BasicVSR++~\cite{chan2022basicvsr++}, respectively.
Although PSD~\cite{9366772} and DCP~\cite{he2010single} get better FADE scores, they generally over-process smoky videos and result in over-saturated colors (as shown in \cref{fig:visualize_comparison_1}), leading to poor NIQE and PI scores, as well as unsatisfactory visual effects. 
Third, with lower computation cost and fewer model parameters, the proposed SelfSVD-S and SelfSVD$^\ast$-S still perform well, which further illustrates the effectiveness of our methods.

\noindent\textbf{Qualitative Comparison.}
Due to space limitation, we only provide qualitative results of some methods with better quantitative scores in \cref{fig:visualize_comparison_1}.
It can be seen that the compared methods often introduce color distortion, over-smoothing, or smoke preservation in their results. 
Instead, our methods remove more smoke and restore more details that are consistent with PS frames. 
More qualitative comparison results can be seen in the Suppl.

\section{Ablation Study}
\subsection{Effect of Taking PS Frame as Supervision}
PS frame (\ie, $\mathbf{S}_{ps}$) is regarded as unaligned supervision in our methods.
To mitigate the adverse effects of misalignment between output and target supervision, we deploy a pre-trained PWC-Net~\cite{sun2018pwc} to align the desmoking result $\mathbf{\hat{I}}_i$ with $\mathbf{S}_{ps}$, then calculate reconstruction loss $\mathcal{L}_{rec}$, as shown in \cref{eq:warpL1}.
Here we conduct experiments with different loss variants to validate the effectiveness of our method. 
The quantitative results are shown in ~\cref{tab:ab_supervision}.
First, we can train the desmoking network without $\mathcal{L}_{rec}$, \ie, only with the adversarial loss.
In this case, the method degenerates to the unpaired learning manner, and it leads to a significant performance drop.
Second, when we do not consider the misalignment issue and utilize the supervision with naive $\ell_1$ loss, it still results in a severe performance drop.
Third, we compare our deformation-based loss with alternative misalignment-tolerated methods, including the contextual bilateral (CoBi) loss~\cite{zhang2019zoom}  and reversed-order alignment (\ie, `align $\mathbf{S}_{ps}$ with $\mathbf{\hat{I}}_i$'). 
Although these two methods can also improve performance compared with naive $\ell_1$ loss, our method achieves better results than theirs.
Compared with the CoBi loss, our method exploits the motion prior in the pre-trained optical flow estimation network, achieving more accurate spatial matching.
Compared with the reversed-order alignment strategy that partially destroys the information of PS frame, our method keeps PS frame unchanged, thus providing more accurate supervision.

\begin{figure}[t!]
\centering
\begin{overpic}[width=0.8\textwidth,grid=False]
{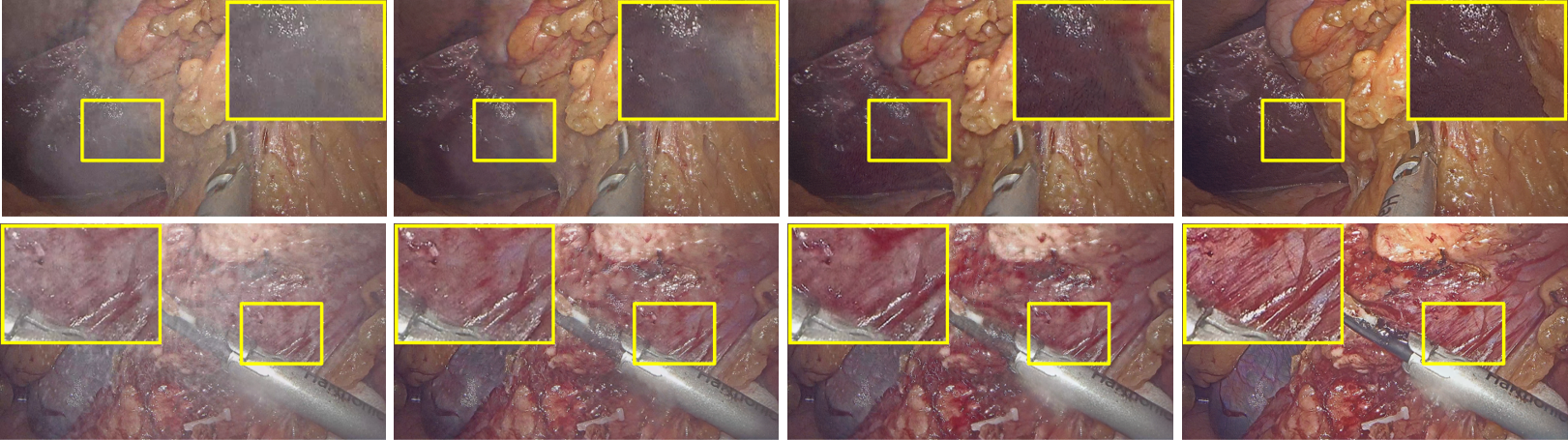}
  \put(5,-8){\scriptsize{(a) Somky Frame}}
  \put(80,-8){\scriptsize{(b) w/o Ref}}
  \put(150,-8){\scriptsize{(c) w/ Ref}}
  \put(220,-8){\scriptsize{(d) PS Frame}}
\end{overpic}
\caption{Qualitative results of taking PS frame as Ref input or not. 
}
\label{fig:ab_reference}
\vspace{-6mm}
\end{figure}
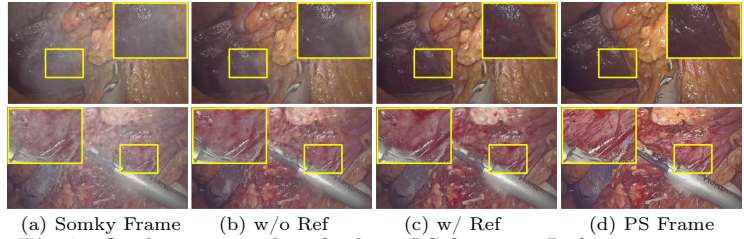

\subsection{Effect of Taking PS Frame as Reference Input}

Note that PS frame is available during training and testing.
We take it as an additional reference (Ref) input.
We perform experiments to evaluate its effectiveness by removing the Ref.
As shown in \cref{tab:reference}, the additional Ref enables 0.67dB PSNR improvements.
Besides, \cref{fig:ab_reference} shows that it leads to cleaner smoke removal and finer-scale detail recovery, especially in areas with dense smoke.

We also conduct experiments to validate whether PS frame is a suitable choice to be used as Ref.
For comparison, we replace Ref with the first smoky frame, as it generally has less smoke than subsequent smoky ones.
As shown in \cref{tab:reference}, the replacement leads to 0.28dB PSNR drop, which illustrates the frames before the activations of high-energy devices are preferable to be taken as Ref.

\subsection{Effect of Strategies to Avoid Trivial Solutions}
\begin{figure*}[t!]
\centering
\begin{overpic}[width=0.99\textwidth,grid=False]
{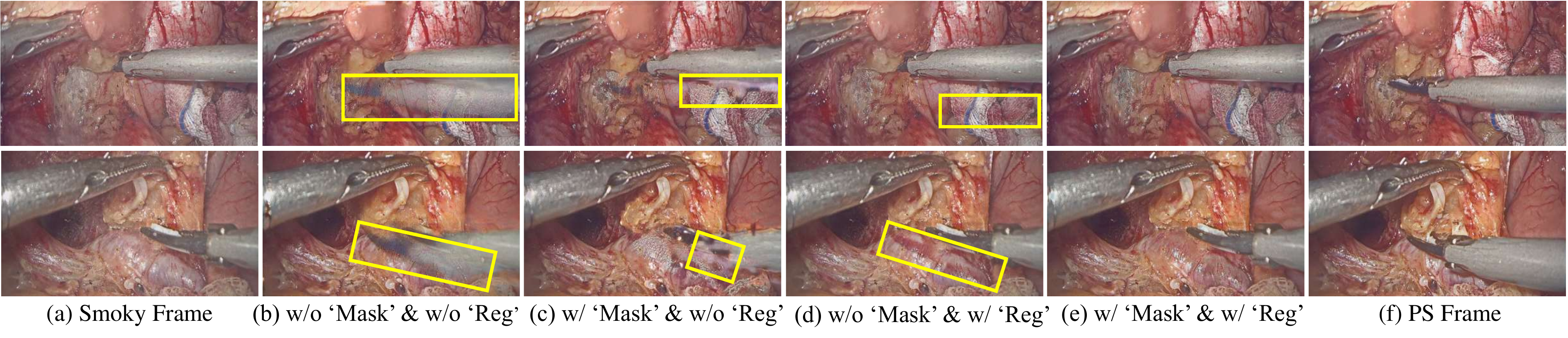}
\end{overpic}
\caption{
Effect of strategies to avoid trivial solutions.
`Mask' and `Reg' denote masking strategy and regularization term, respectively.
Naively inputting PS frame as Ref leads to trivial solutions, as marked with yellow boxes in (b).
Using `Mask' alone may generate artifacts around high-energy devices, as marked with yellow boxes in (c).
Using `Reg' alone may leave trivial solution traces of high-energy devices from Ref, as marked with yellow boxes in (d).
Their combination produces better results in (e).
} 
\label{fig:ab_overfitting}
\end{figure*}

A masking strategy and a regularization term are introduced to prevent trivial solutions.
To validate the effectiveness, we conduct experiments with their different combinations, \ie, `w/o Mask \& w/o Reg', `w/ Mask \& w/o Reg', `w/o Mask \& w/ Reg' and `w/ Mask \& w/ Reg'.
`Mask' and `Reg' denote the masking strategy and the regularization term respectively.
Naively inputting PS frame as Ref easily leads to trivial solutions, as marked with yellow boxes in \cref{fig:ab_overfitting} (b).
\cref{fig:ab_overfitting} (c) and (d) show that both `Mask' and `Reg' inhibit the trivial solutions, but using one of them alone does not achieve the best results.
On the one hand, using `Mask' alone can handle poorly aligned areas well, as they are relatively easy to detect. 
Nevertheless, there still remain artifacts around high-energy devices, as marked with yellow boxes in \cref{fig:ab_overfitting} (c).
On the other hand, using `Reg' alone generally suppresses trivial solutions.
As the areas with significantly imperfect optical flow are not explicitly processed, it may leave some traces of high-energy devices from Ref, as marked with yellow boxes in \cref{fig:ab_overfitting} (d).
Instead, their combination generates improved results, as shown in \cref{fig:ab_overfitting} (e).
Besides, we also provide the effect of regularization loss weight $\lambda_{reg}$ in the Suppl.

\begin{table}[t!]
\centering 
\begin{minipage}[c]{0.48\linewidth} 
\centering
\caption{Ablation studies on reconstruction loss $\mathcal{L}_{rec}$. 
}
\vspace{-2mm}
\scalebox{0.7}{ 
\begin{tabular}{cc}
\toprule
$\mathcal{L}_{rec}$ & PSNR$\uparrow$/SSIM$\uparrow$/FADE$\downarrow$/NIQE$\downarrow$/PI$\downarrow$ \\
\midrule
None & 22.40/0.6030/0.4219/5.37/4.04 \\
Naive $\ell_1$ Loss & 22.67/0.5880/0.4720/5.23/4.00 \\
CoBi Loss~\cite{zhang2019zoom} & 22.82/0.6093/0.4721/4.86/3.86 \\
Align $\mathbf{S}_{ps}$ with $\mathbf{\hat{I}}_i$  & 24.12/0.6205/0.4663/5.05/3.94\\
Align $\mathbf{\hat{I}}_i$ with $\mathbf{S}_{ps}$ & 24.23/0.6225/0.4626/4.85/3.87 \\
\bottomrule
\end{tabular}
}
\label{tab:ab_supervision}
\end{minipage}
\hfill 
\begin{minipage}[c]{0.48\linewidth}
\centering
\caption{Quantitative results when inputting different Ref.
`None' denotes no Ref being input.
}
\vspace{-2mm}
\scalebox{0.7}{ 
\begin{tabular}{cc}
\toprule
Input Ref & PSNR$\uparrow$/SSIM$\uparrow$/FADE$\downarrow$/NIQE$\downarrow$/PI$\downarrow$\\
\midrule
None & 23.56/0.6183/0.4772/4.87/3.85 \\
First Smoky Frame  & 23.95/0.6216/0.4701/4.88/3.85 \\
PS Frame & 24.23/0.6225/0.4626/4.85/3.87 \\
\bottomrule
\end{tabular}
}
\label{tab:reference}
\end{minipage}
\vspace{-4mm}
\end{table}

\section{Conclusion}
Existing laparoscopic surgery desmoking works struggle with processing real-world smoke, especially dense smoke, due to the unavailable of real-world paired data and the severely ill-posed nature of single-image methods.
To address the issue, we suggested leveraging the internal characteristics of real-surgery video for effective self-supervised video desmoking and propose SelfSVD to achieve this. 
Based on the observation that the pre-smoke (PS) frame has less smoke and similar contents as subsequent smoky ones, SelfSVD utilizes it as an unaligned supervision.
Moreover, SelfSVD takes PS frame as a reference input to handle dense smoke better, and a masking strategy and a regularization term are introduced to prevent trivial solutions.  
Besides, we collect a real-world laparoscopic surgery video desmoking (LSVD) dataset, which can potentially be advantageous for future studies.
Extensive experiments show that our SelfSVD outperforms the state-of-the-art methods both quantitatively and qualitatively.

\section*{Acknowledgement}
This work was supported in part by the National Natural Science Foundation of China (NSFC) under Grant No. 62371164 and No. U22B2035.

\clearpage


\title{Self-Supervised Video Desmoking for Laparoscopic Surgery \\ (Supplementary Material)}   
\titlerunning{SelfSVD}

\renewcommand{\thesection}{\Alph{section}}
\renewcommand{\thetable}{\Alph{table}}
\renewcommand{\thefigure}{\Alph{figure}}
\renewcommand{\thealgorithm}{\Alph{algorithm}}
\renewcommand{\theequation}{\Alph{equation}}

\author{Renlong Wu\inst{1} \and
Zhilu Zhang\inst{1}$^{(}$\Envelope$^)$ 
\and
Shuohao Zhang\inst{1}
\and
Longfei Gou\inst{2}
\and
\\
Haobin Chen\inst{2}
\and
Lei Zhang \inst{3}
\and
Hao Chen \inst{2}$^{(}$\Envelope$^)$ 
\and
Wangmeng Zuo \inst{1}
}

\authorrunning{R.Wu et al.}

\institute{
Harbin Institute of Technology, China \and
Southern Medical University, China \and
Hong Kong Polytechnic University, China
\\
\email{hirenlongwu@gmail.com,cszlzhang@outlook.com,
\\yhyzshrby@163.com,Calvin\_smu@163.com,Hao\-Bin\_Chen@outlook.com,
\\cslzhang@comp.polyu.edu.hk,chenhao.05@163.com,wmzuo@hit.edu.cn}
}

\maketitle

\noindent{The content of the supplementary material involves:}
\begin{itemize}
\item Results on synthetic dataset in Sec.~\ref{sec:supp_0}.
\vspace{2mm}
\item Networks details in Sec.~\ref{sec:supp_2}.
\vspace{2mm}
\item Visualization of mask in Sec.~\ref{sec:supp_3}.
\vspace{2mm}
\item Effect of regularization term  in Sec.~\ref{sec:supp_4}.
\vspace{2mm}
\item Examples from LSVD dataset  in Sec.~\ref{sec:supp_5}.
\vspace{2mm}

\item More result comparisons in Sec.~\ref{sec:supp_6}.
\vspace{2mm}
\item Practical deployment in surgery in Sec.~\ref{sec:supp_1}.
\vspace{2mm}
\item Limitation and social impact in Sec.~\ref{sec:supp_7}.

\end{itemize}

\section{Results on Synthetic Dataset}
\label{sec:supp_0}
\begin{figure*}[t!]
\begin{overpic}[width=0.99\textwidth,grid=False]
{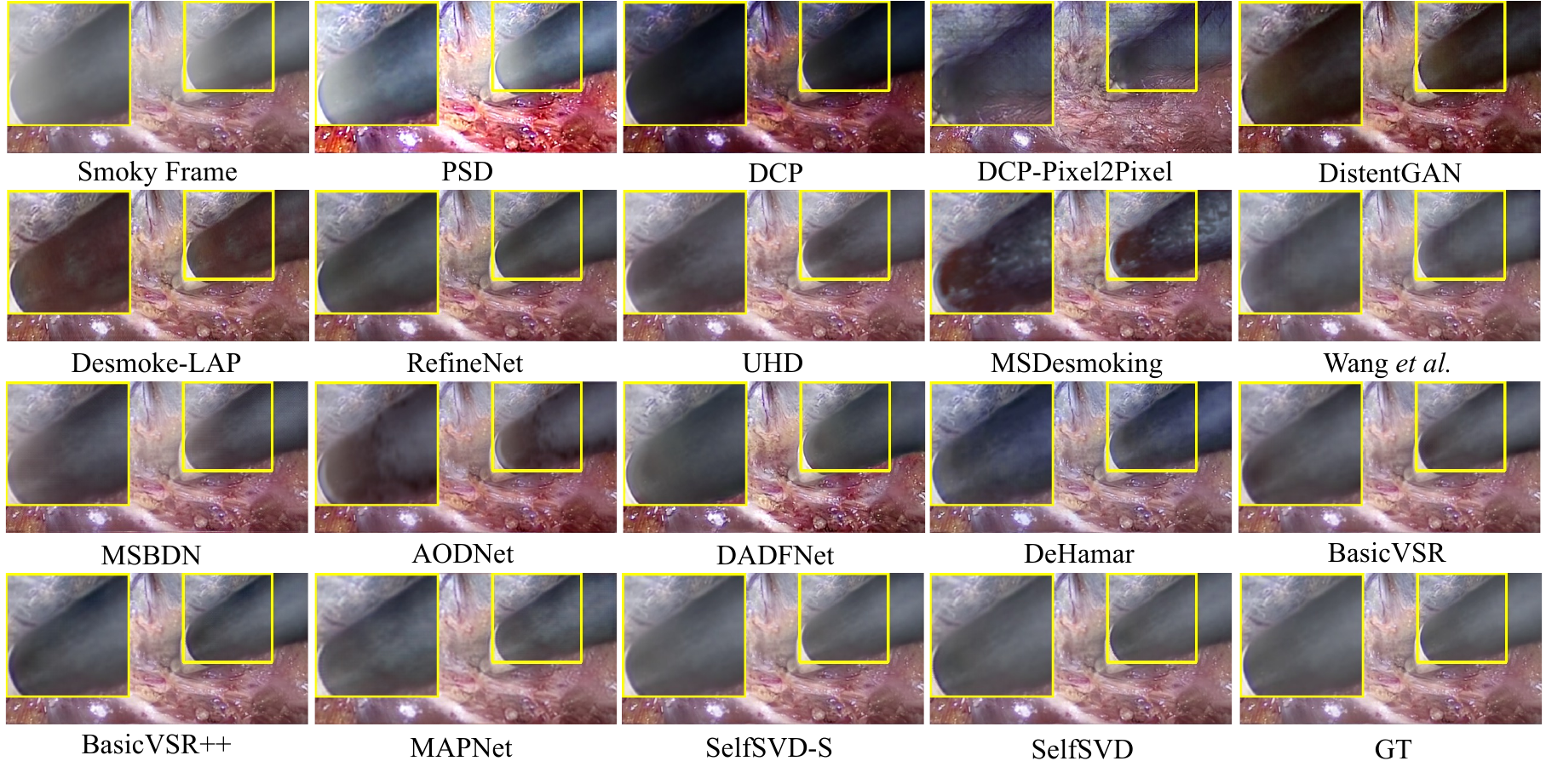}
\end{overpic}
\caption{Qualitative comparisons on the synthetic dataset. Our results produce few visual artifacts and are more consistent with the GT.}
\label{fig:syn_vis_1}
\end{figure*}

\begin{figure*}[t!]
\begin{overpic}[width=0.99\textwidth,grid=False]
{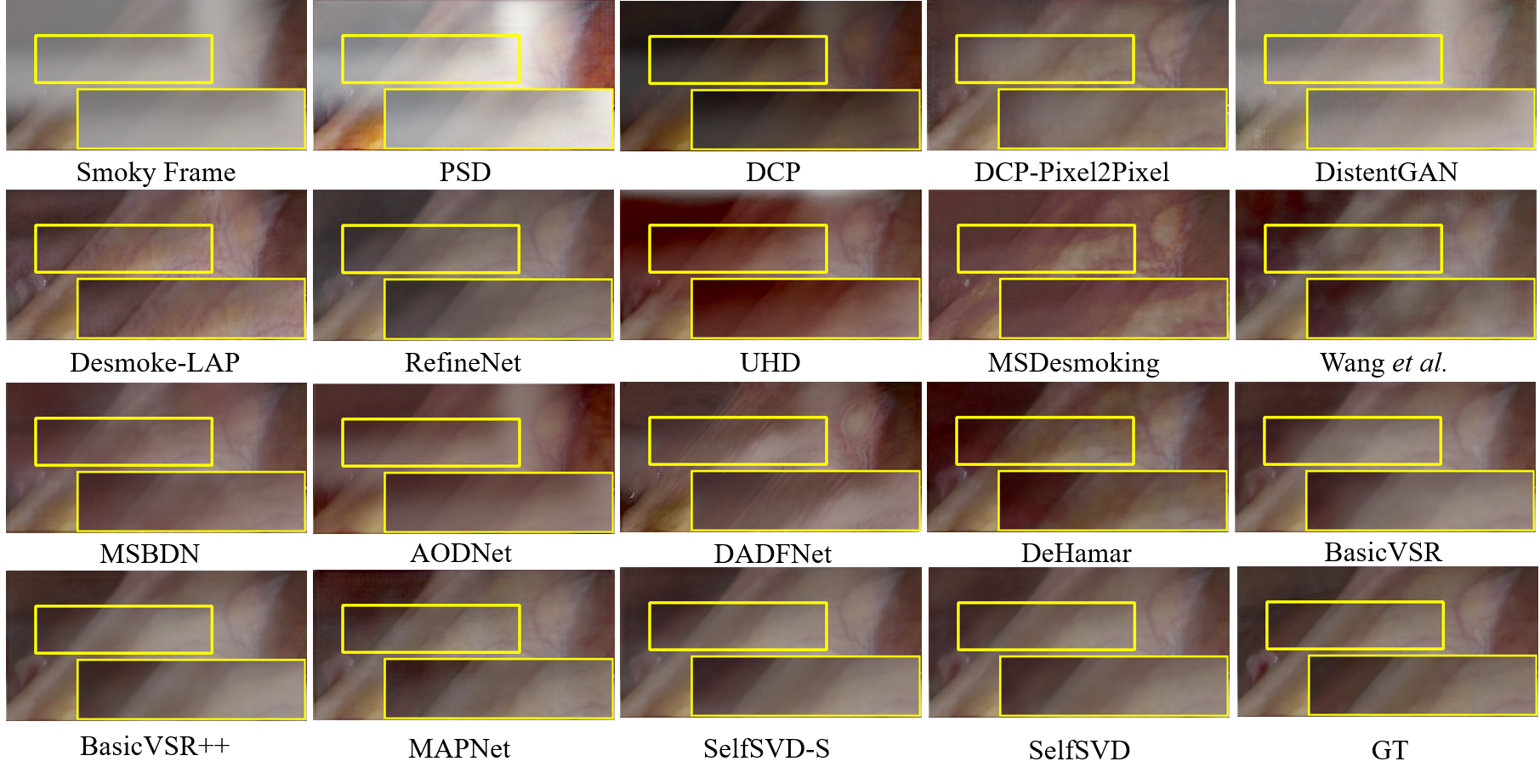}
\end{overpic}
\caption{Qualitative comparisons on the synthetic dataset. Our results remove more clean smoke and are more consistent with the GT.}
\label{fig:syn_vis_2}
\end{figure*}

We also conduct experiments in a synthetic video dataset.
We evaluate the compared methods with paired clean videos that are spatially aligned with the smoky videos.
We collect 380 clean video clips from public Cholec80 dataset~\cite{twinanda2016endonet}, where 280 clips are used for training and the remaining 100 ones are used for evaluation.
We follow the surgery smoke simulation manner~\cite{hong2023mars} and add synthetic smoke in the clean video clips, except the first frame (regarded as PS frame).
The results are shown in \cref{tab:syn_comparison_supp}.
As the PS frame in the synthetic dataset is clean, we do not perform experiments with SelfSVD$^\ast$ and SelfSVD$^\ast$-S. 
Our results get the best PSNR scores, indicating the effectiveness of the proposed method.
Moreover, the visual comparisons in \cref{fig:syn_vis_1,fig:syn_vis_2} show that our results produce few visual artifacts, remove more clean smoke and are more consistent with the GT.

\begin{table*}[t!] 
    \small
    \setlength{\tabcolsep}{5mm}
    \renewcommand\arraystretch{1}
    \begin{center}
	\caption{
 Comparison results on the synthetic dataset.
 The best results in each category are marked in \pmb{bold}.
 }
	\label{tab:syn_comparison_supp}
	\begin{tabular}{clccccc}
		\toprule
		 & Methods & PSNR$\uparrow$ & SSIM$\uparrow$ \\
        \midrule
        \multirow{2}{*}{\begin{tabular}[c]{@{}c@{}} Unsupervised \\ Image Processing  \end{tabular}} & PSD~\cite{chen2021psd} & 13.94 & \pmb{0.7683}  \\
        \multicolumn{1}{c}{} & DCP~\cite{he2010single} & \pmb{15.71} & 0.7413 \\
        \midrule
        \multirow{4}{*}{\begin{tabular}[c]{@{}c@{}} Unpaired \\ Image Processing  \end{tabular}} 
        & DCP-Pixel2Pixel~\cite{salazarcolores2020desmoking} & 20.01 & 0.5226 \\
        \multicolumn{1}{c}{} 
        & DistentGAN~\cite{shyam2021towards} & 20.88 & 0.8056 \\
        \multicolumn{1}{c}{} 
                & Desmoke-LAP~\cite{pan2022desmoke} & 24.82 & 0.8966 \\ 
        \multicolumn{1}{c}{} 
                & RefineNet~\cite{9366772} & \pmb{26.94} & \pmb{0.9334} \\ 
        \midrule
        \multirow{7}{*}{\begin{tabular}[c]{@{}c@{}} Self-Supervised \\ Image Processing  \end{tabular}} & UHD~\cite{xiao2022single} & 25.50 & 0.9410\\
        \multicolumn{1}{c}{} 
                & MSDesmoking~\cite{wang2019multiscale} & 26.77 & 0.9021\\
        \multicolumn{1}{c}{}
                & Wang \etal~\cite{wang2023surgical} & \pmb{30.76} & \pmb{0.9470} \\  
        \multicolumn{1}{c}{} 
                & MSBDN~\cite{dong2020multi} & 23.27 & 0.6251 \\ 
        \multicolumn{1}{c}{}
                & AODNet~\cite{li2017aod} & 22.69 & 0.9002 \\
        \multicolumn{1}{c}{} 
                & DADFNet~\cite{guo2023dadfnet} & 23.32 & 0.8264 \\
        \multicolumn{1}{c}{} 
                & DeHamar~\cite{guo2022image} & 29.06 & 0.9396\\
        \midrule
        \multirow{5}{*}{\begin{tabular}[c]{@{}c@{}} Self-Supervised \\ Video Processing  \end{tabular}} 
                & BasicVSR~\cite{chan2021basicvsr} & 31.33 & 0.9503 \\
        \multicolumn{1}{c}{} 
                & BasicVSR++~\cite{chan2022basicvsr++} & 31.84 & 0.9570\\
        \multicolumn{1}{c}{} 
                & MAPNet~\cite{xu2023video} & 31.29 & 0.9242\\
                & (Ours) SelfSVD-S & 31.96 & 0.9520\\
        \multicolumn{1}{c}{} 
                & (Ours) SelfSVD & \pmb{32.30} & \pmb{0.9611}\\
		\bottomrule
	\end{tabular}
    \end{center}
\end{table*}

\section{Network Details}
\label{sec:supp_2}
We design the video desmoking network based on the unidirectional recurrent network~\cite{wu2023rbsr}.
It includes five modules, \ie, feature encoder, masked-ref generator, alignment, fusion, and reconstruction module.
When processing the $i$-th smoky frame $\mathbf{S}_{i}$, it is first fed into the encoder to obtain feature representations $\mathbf{F}_{i}$.
$\mathbf{S}_{ref}$ and $\mathbf{S}_{i}$ are fed into the masked-ref generator to get the masked reference features $\tilde{\mathbf{F}}_{ref\rightarrow i}$.
Then, we deploy the alignment module to align previous temporal features $\mathbf{H}_{i-1}$ to $\mathbf{F}_{i}$, getting the warped ones $\mathbf{H}_{i-1 \rightarrow i}$.
Next, the fusion module concatenates $\mathbf{F}_{i}$, $\tilde{\mathbf{F}}_{ref\rightarrow i}$ and $\mathbf{H}_{i-1 \rightarrow i}$ as inputs, producing the fused features.
Finally, the fused features are fed into the reconstruction module to generate the restored clean component $\mathbf{\hat{I}}_{i}$.

\noindent\textbf{Details of SelfSVD and SelfSVD$^\ast$.}
For SelfSVD and SelfSVD$^\ast$, the encoder includes two $3\times3$ convolutional layers with a stride of 2 for scale down-sampling and 5 residual blocks~\cite{he2016deep} for feature extraction.
We deploy individual encoders for the smoky frame and reference input respectively.
The alignment module is built upon a pre-trained optical flow network (\eg, PWC-Net~\cite{sun2018pwc}).
The fusion module consists of a $3\times3$ convolutional layer for channel reduction and 60 residual blocks for feature enhancement.
The reconstruction module includes 5 residual blocks, two pixel-shuffle operations, and a final $3 \times 3$ convolutional layer.

\begin{figure}[t!]
    \centering
    \includegraphics[width=0.98\linewidth]{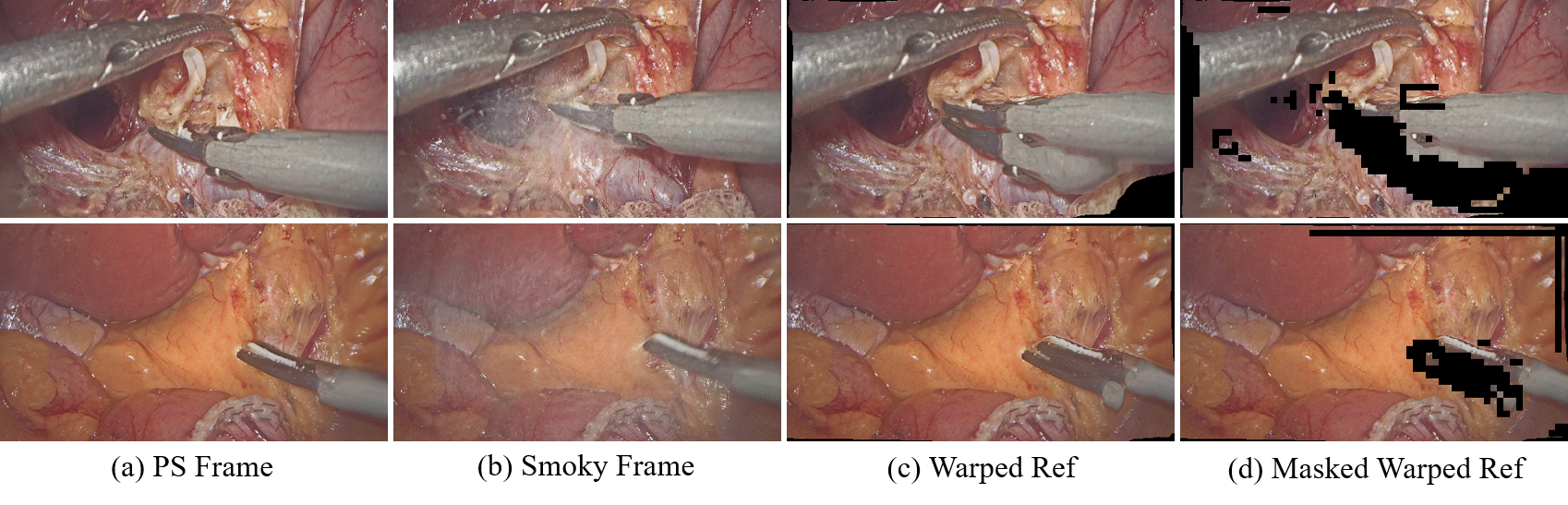}
    \caption{Mask examples. 
    A mask is generated to indicate areas with imperfect optical flow to prevent trivial solutions. 
    Masked regions are marked with black pixels in (d). 
    }
    \label{fig:mask_samples}
\end{figure}

\begin{table}[t!]
\centering
\caption{Structure configuration of the discriminator. The kernel size of all convolutional layers is $4\times4$. `Stride' denotes the stride of convolutional layer. `BN' denotes the BatchNorm~\cite{ioffe2015batch} operation.}
\scalebox{1.2}{\begin{tabular}{p{4cm}<{\centering} p{2cm}<{\centering} p{2cm}<{\centering} p{2cm}<{\centering}}
\toprule
 Layer & Filter & Stride &Output size \\
\midrule
 Conv, LeakyReLU & $3 \rightarrow 64$ & 2 & $128 \times 128$ \\
 Conv, BN, LeakyReLU & $64\rightarrow 128$ & 2& $64 \times 64$ \\
 Conv, BN, LeakyReLU & $128\rightarrow 256$ & 2& $32 \times 32$ \\
 Conv, BN, LeakyReLU & $256\rightarrow 512$ & 2& $31 \times 31$ \\
 Conv, BN, LeakyReLU & $512\rightarrow 1$ & 1 & $30 \times 30$ \\
\toprule
\end{tabular}}
\label{tab:patch_gan}
\end{table}

\noindent\textbf{Details of SelfSVD-S and SelfSVD$^\ast$-S.}
The computation costs of SelfSVD and SelfSVD$^\ast$ are generally similar to the video processing methods.
To make the computation cost consistent with single-image ones, we present the lightweight models SelfSVD-S and SelfSVD$^\ast$-S.
Specifically, we first replace  
PWC-Net~\cite{sun2018pwc} in the alignment module to  a more lightweight optical flow network SpyNet~\cite{ranjan2017optical}.
Second, we reduce the number of residual blocks in the feature encoder, masked-ref generator, fusion, and reconstruction module from 5, 5, 60, 5 to 3, 3, 8, 3 respectively. 
Third, we reduce the channel numbers from 64 to 32.
Benefiting from the above simplification, SelfSVD-S and SelfSVD$^\ast$-S significantly reduce the number of model parameters and computation costs, while keeping performance.
\noindent\textbf{Details of Discriminator.} PatchGAN~\cite{zhu2017unpaired} is employed as the discriminator to distinguish whether a patch is real or fake. 
Its structure is shown in \cref{tab:patch_gan}.

\section{Visualization of Mask}
\label{sec:supp_3}
\begin{figure}[t!]
    \centering
    \includegraphics[width=0.98\linewidth]{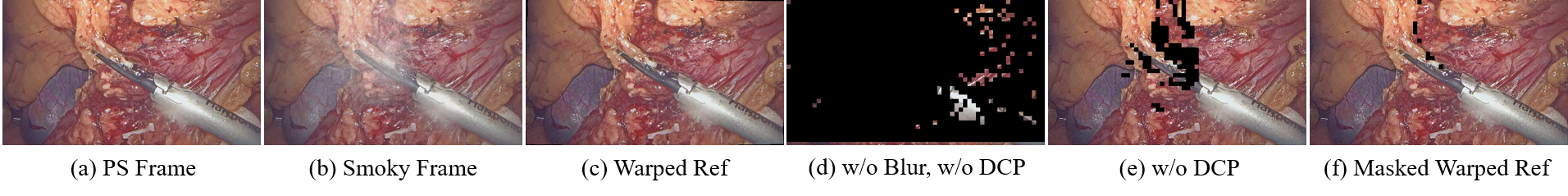}
    \caption{Effect of dark channel prior (DCP) and Gaussian blur (Blur) operations.  
    DCP and Blur help to avoid detecting incorrect masked areas. 
    The masked areas are marked with black pixels in (d), (e), and (f).}
    \label{fig:mask_samples_2}
\end{figure}

To prevent trivial solutions, we generate a mask to indicate areas with imperfect optical flow.
Moreover, we process the smoky frame and warped reference (Ref) input with dark channel prior~\cite{he2010single} (DCP) and large-kernel Gaussian blur (Blur) to mitigate smoke interference.
Mask examples are provided in \cref{fig:mask_samples,fig:mask_samples_2}.
On the one hand,, it successfully detects the areas with imperfect optical flow, as shown in  \cref{fig:mask_samples}.
On the other hand, DCP and Blur help to avoid detecting incorrect masked areas, as shown in  \cref{fig:mask_samples_2}.

\section{Effect of Regularization Term}
\label{sec:supp_4}
\begin{figure}[t!]
\centering
\begin{overpic}[width=0.98\textwidth,grid=False]
{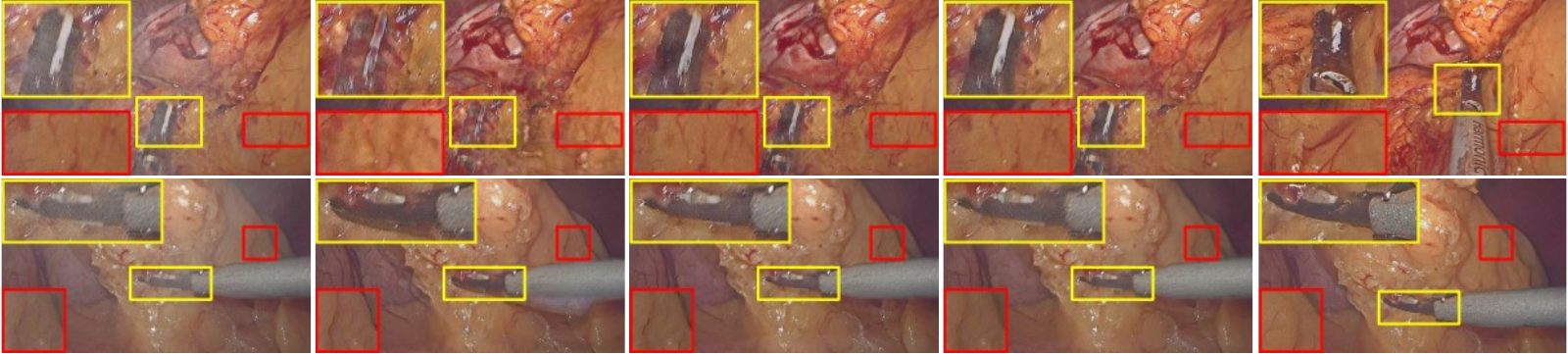}
  \put(10,-8){\scriptsize{Somky Frame}}
  \put(80,-8){\scriptsize{$\lambda_{reg}=0.01$}}
  \put(150,-8){\scriptsize{$\lambda_{reg} = 0.05$}}
  \put(220,-8){\scriptsize{$\lambda_{reg} = 0.1$}}
  \put(290,-8){\scriptsize{PS Frame}}
\end{overpic}
\caption{Effect of regularization term weight $\lambda_{reg}$. 
Please zoom in for better observation.
}
\label{fig:ab_lambda}
\end{figure}
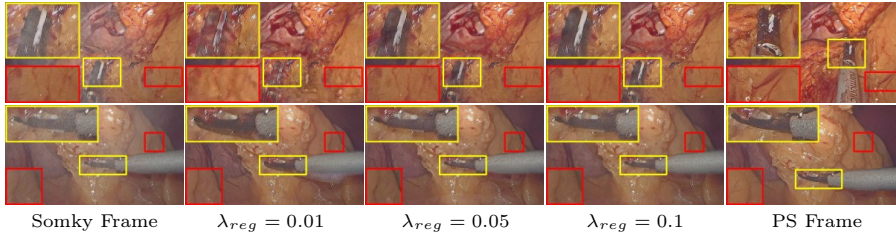

Here we conduct experiments to validate the effect of regularization term weight $\lambda_{reg}$.
The results are shown in \cref{fig:ab_lambda} and \cref{tab:ab_lambda_weight}.
In general, a smaller $\lambda_{reg}$ leads to a weaker suppression of trivial solutions, while a larger one leads to a weaker utilization of Ref.
We set $\lambda_{reg}$ to 0.05 for the trade-off.

\section{Examples from LSVD Dataset}
\label{sec:supp_5}
\begin{figure*}[t!]
    \centering
    \includegraphics[width=0.99\linewidth]{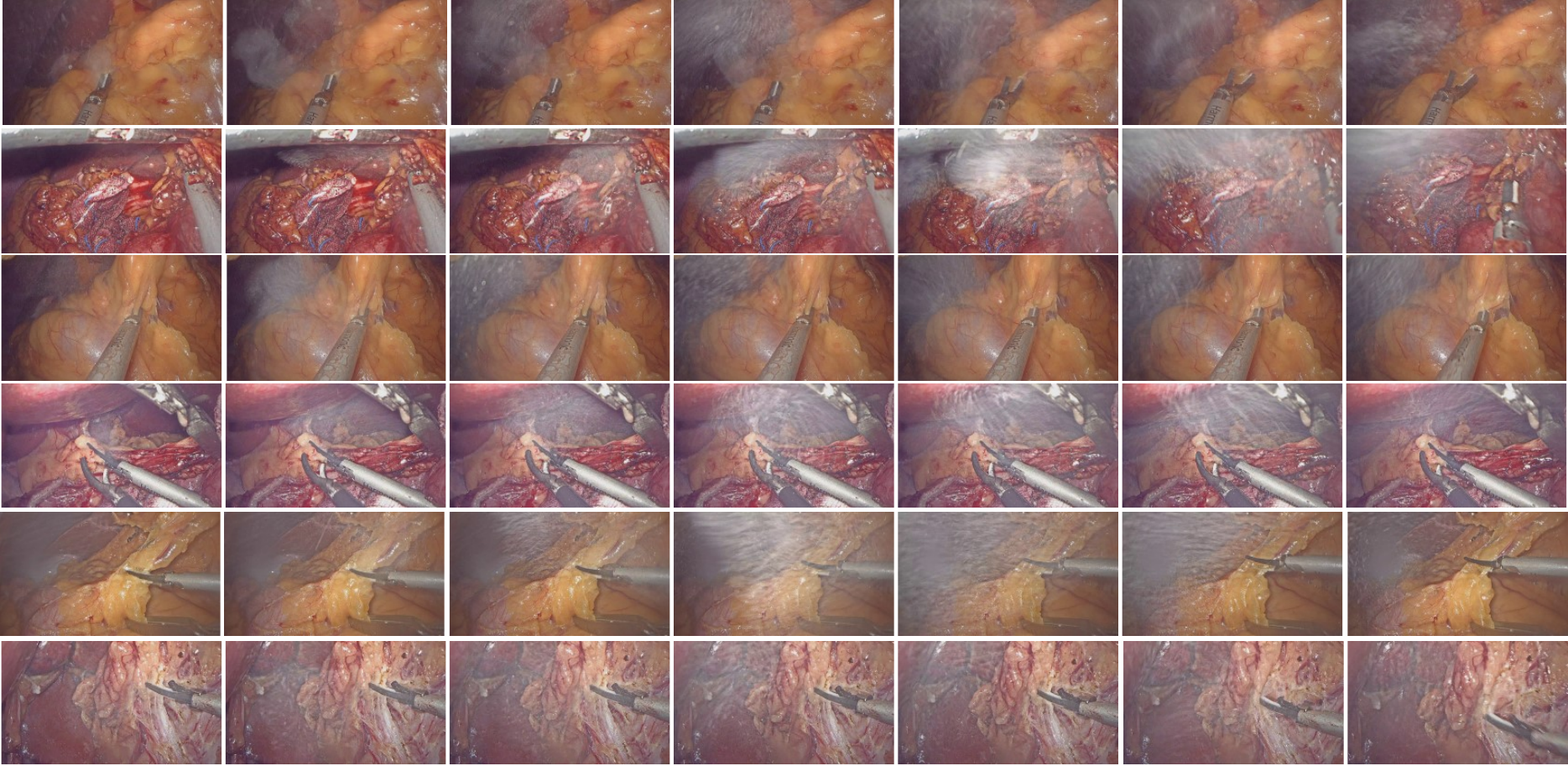}
    \caption{Examples from LSVD dataset. 
    Each row shows several frames from a video clip.
    It can be seen that the dataset contains diverse and complex surgery smoke, such as mist, droplets, and streaks.
    Please zoom in for better observation.}
    \label{fig:dataset_samples}
\end{figure*}

We construct the laparoscopic surgery video desmoking (LSVD) dataset from professional hospitals. 
Some examples from the dataset are provided in \cref{fig:dataset_samples}.
It can be seen that the dataset contains diverse and complex surgery smoke, such as mist, droplets, and streaks.

\begin{table}[t!]
\centering
\caption{Effect of regularization term weight $\lambda_{reg}$. }
\scalebox{1.2}{\begin{tabular}{p{2cm}<{\centering} p{8cm}<{\centering}}
\toprule
 $\lambda_{reg}$ & PSNR$\uparrow$ / SSIM$\uparrow$ / FADE$\downarrow$ / NIQE$\downarrow$ / PI$\downarrow$\\
\midrule
 $0.01$  & 24.46 / 0.6254 / 0.4650 / 4.91 / 3.88\\
 $0.05$  & 24.23 / 0.6225 / 0.4626 / 4.85 / 3.87\\
 $0.1$  & 24.03 / 0.6200 / 0.4658 / 4.87 / 3.82\\
\bottomrule
\end{tabular}}
\label{tab:ab_lambda_weight}
\end{table}

\section{More Result Comparisons}
\label{sec:supp_6}

In the main text, we utilize the processed PS frame as the target to evaluate desmoking methods, as smoke may remain in it.
Here, we provide the comparison results when taking the original PS frame as the target, as shown in \cref{tab:sota_comparison_supp}.
Due to interference from smoke in the PS frame, SelfSVD$^\ast$ and SelfSVD$^\ast$-S get lower PSNR and SSIM values than SelfSVD and SelfSVD-S, respectively.
Moreover, it shows that the proposed methods still outperform state-of-the-art ones.

\begin{figure*}[htp!]
\centering
\begin{overpic}[width=0.99\textwidth,grid=False]
{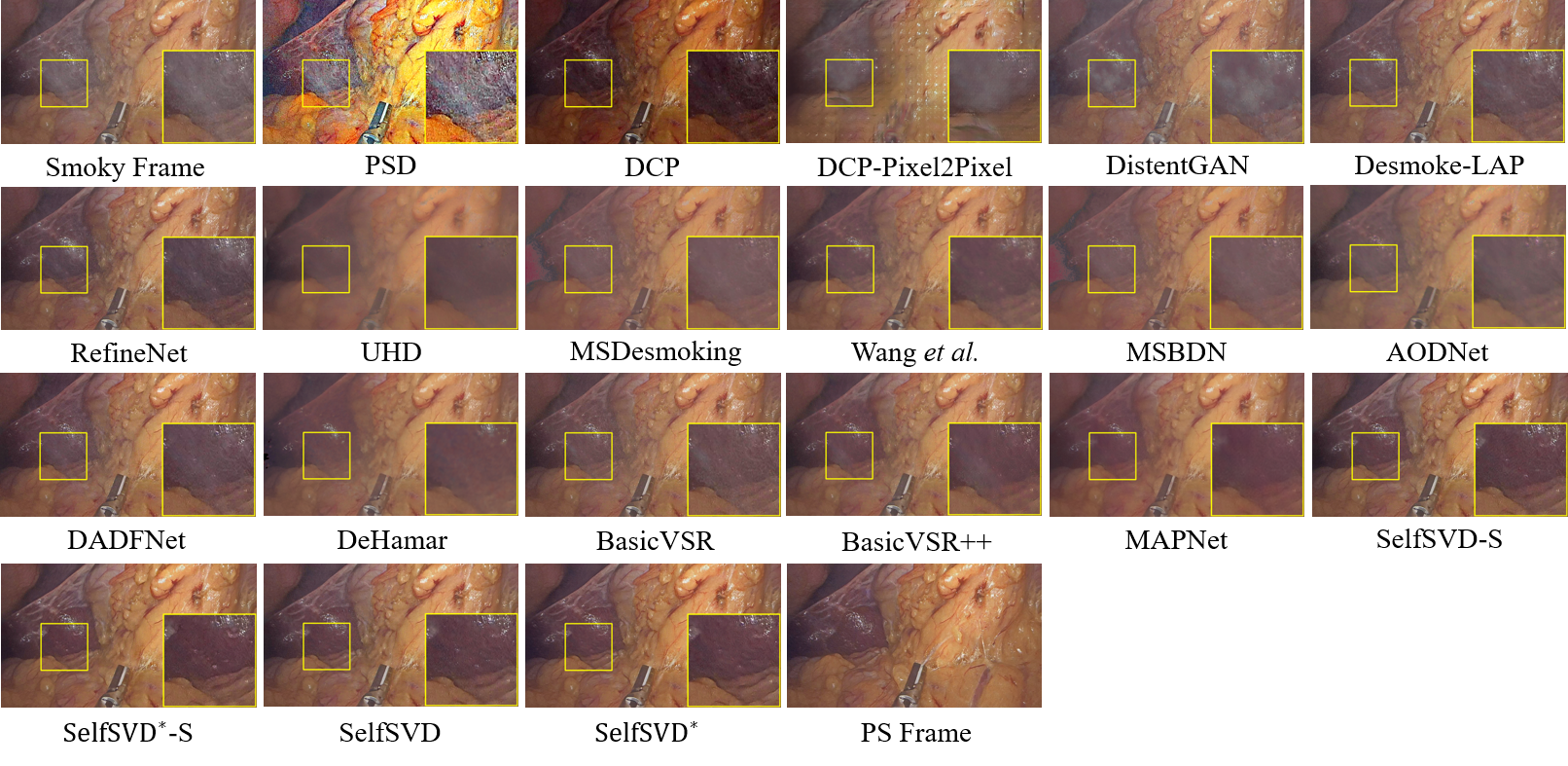}
\end{overpic}
\caption{Qualitative comparisons on LSVD dataset. Our methods generate results more consistent with the PS frame. Please zoom in for better observation.}
\label{fig:more_visualize_comparison_1}
\end{figure*}

\begin{figure*}[htp!]
\begin{overpic}[width=0.99\textwidth,grid=False]
{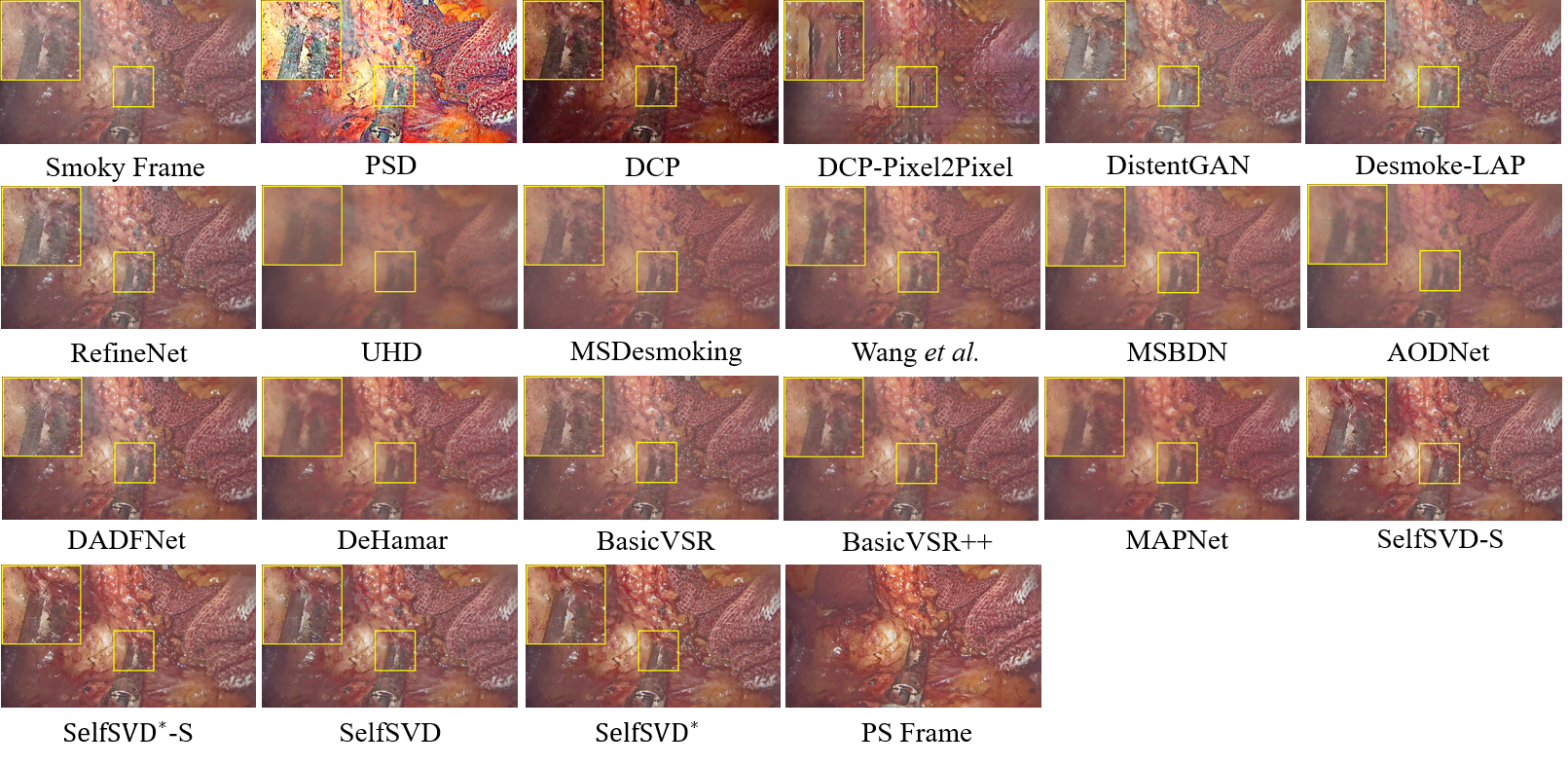}
\end{overpic}
\caption{Qualitative comparisons on LSVD dataset. Our methods remove more clean smoke and recover more realistic details. Please zoom in for better observation.}
\label{fig:more_visualize_comparison_2}
\end{figure*}

\begin{figure*}[htp!]
\begin{overpic}[width=0.99\textwidth,grid=False]
{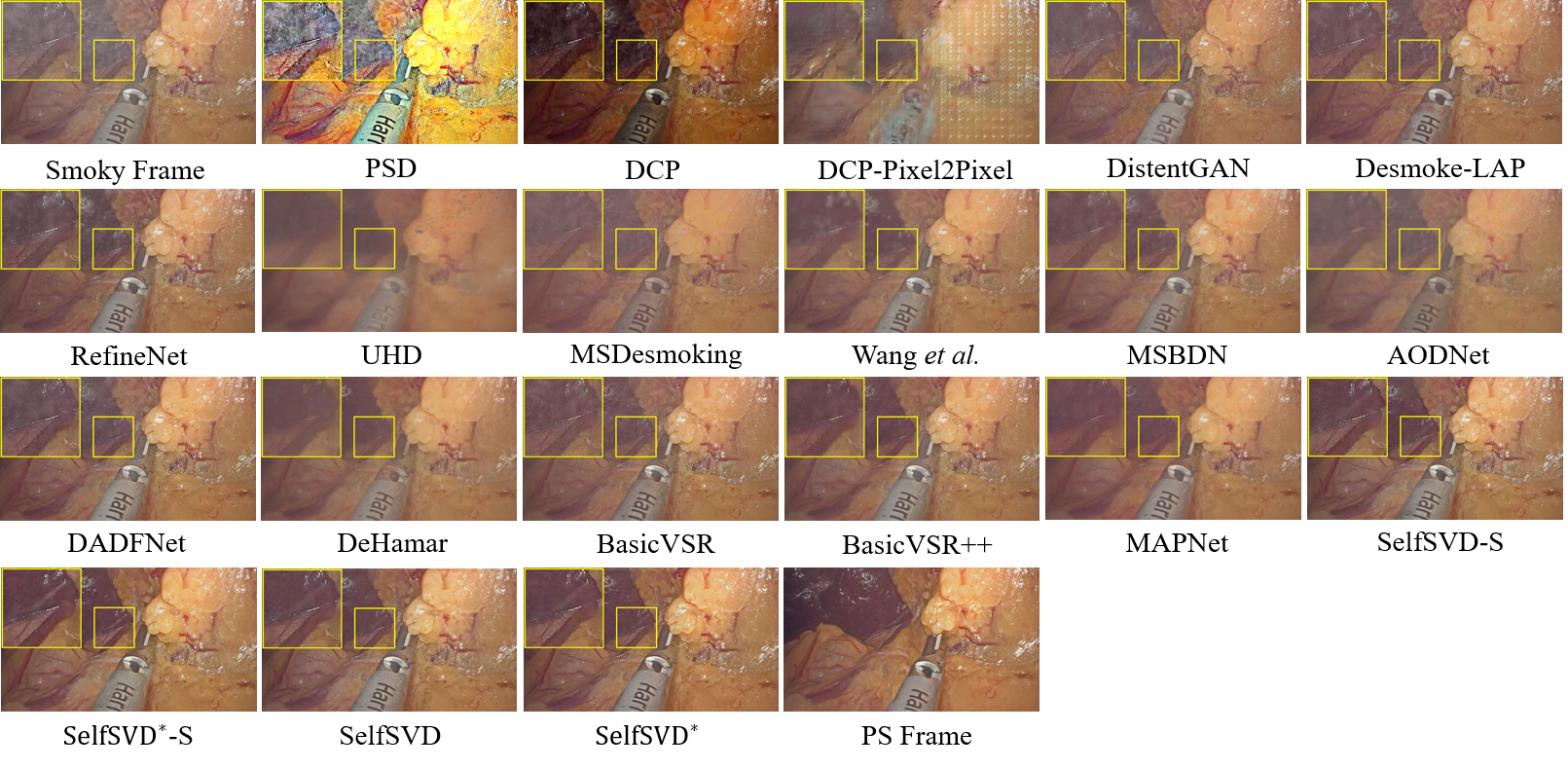}
\end{overpic}
\caption{Qualitative comparisons on LSVD dataset.
Our methods generate results with fewer artifacts and are more consistent with the PS frame. Please zoom in for better observation.}
\label{fig:more_visualize_comparison_3}
\end{figure*}

\begin{figure*}[htp!]
\begin{overpic}[width=0.99\textwidth,grid=False]
{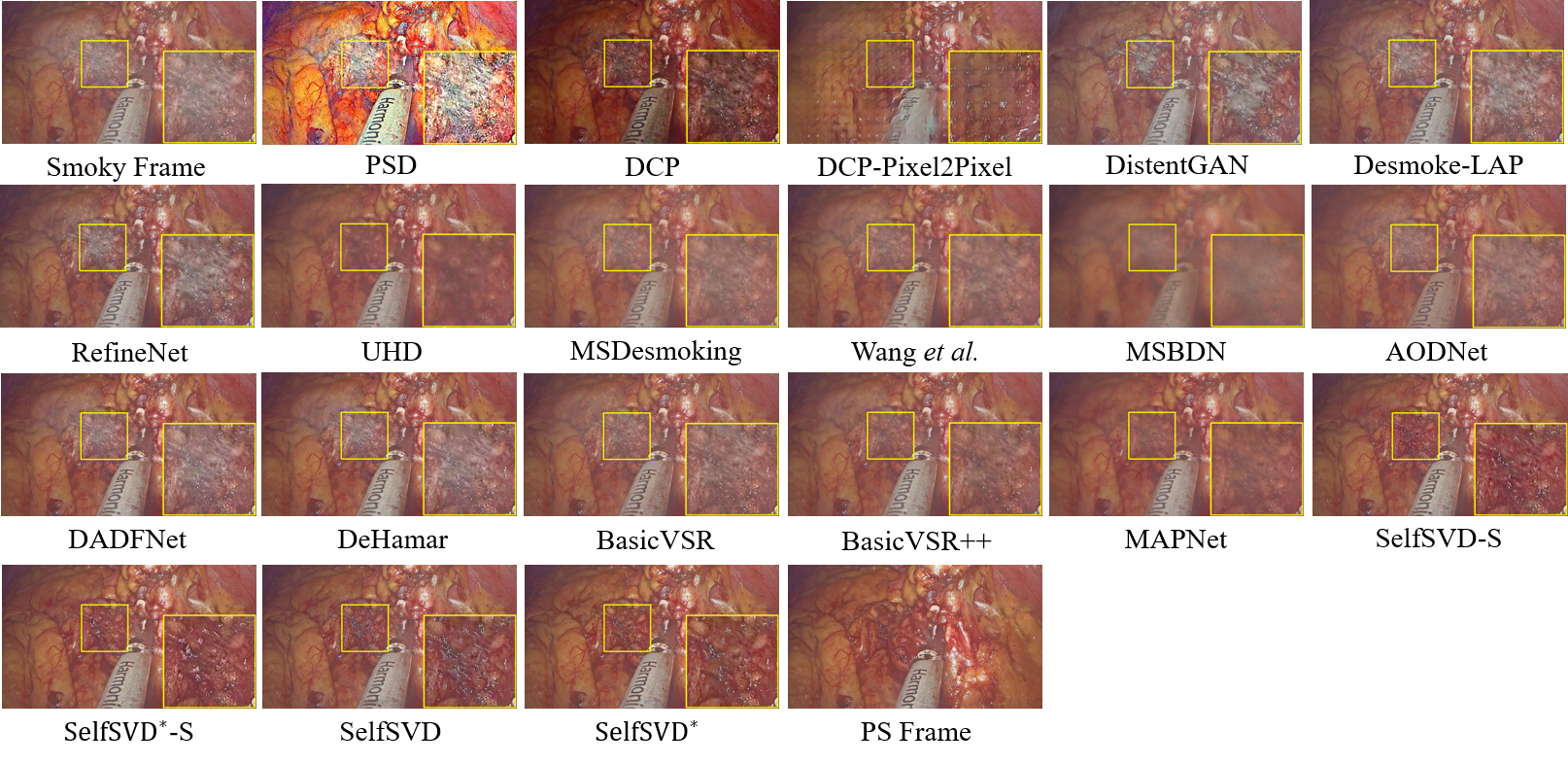}
\end{overpic}
\caption{Qualitative comparisons on LSVD dataset. Our methods remove more clean smoke and recover more fine-scale details. Please zoom in for better observation.}
\label{fig:more_visualize_comparison_4}
\end{figure*}

More qualitative comparison results are provided in \cref{fig:more_visualize_comparison_1,fig:more_visualize_comparison_2,fig:more_visualize_comparison_3,fig:more_visualize_comparison_4}.
Our methods can remove more smoke, recover more photo-realistic details, and produce results more consistent with PS frames than state-of-the-art ones.

\begin{table*}[t!] 
    \small
    \setlength{\tabcolsep}{5mm}
    \renewcommand\arraystretch{1}
    \begin{center}
	\caption{
 Comparisons on LSVD dataset. 
 The original PS frame is taken as the clean target for calculating metrics.
 The best results in each category are marked in \pmb{bold}.
 }
	\label{tab:sota_comparison_supp}
	\begin{tabular}{clccc}
		\toprule
		 & Methods & PSNR$\uparrow$ & SSIM$\uparrow$ \\
        \midrule
        \multirow{2}{*}{\begin{tabular}[c]{@{}c@{}} Unsupervised \\ Image Processing  \end{tabular}} & PSD~\cite{chen2021psd} & 14.49 & 0.3811  \\
        \multicolumn{1}{c}{} & DCP~\cite{he2010single} & \pmb{17.54} & \pmb{0.5819} \\
        \midrule
        \multirow{4}{*}{\begin{tabular}[c]{@{}c@{}} Unpaired \\ Image Processing  \end{tabular}} 
        & DCP-Pixel2Pixel~\cite{salazarcolores2020desmoking} & 20.31 & 0.5409  \\
        \multicolumn{1}{c}{} 
        & DistentGAN~\cite{shyam2021towards} & 22.58 & 0.6429  \\
        \multicolumn{1}{c}{} 
                & Desmoke-LAP~\cite{pan2022desmoke} & 23.70 & 0.6548 \\ 
        \multicolumn{1}{c}{} 
                & RefineNet~\cite{9366772} & \pmb{24.06} & \pmb{0.6559}  \\ 
        \midrule
        \multirow{7}{*}{\begin{tabular}[c]{@{}c@{}} Self-Supervised \\ Image Processing  \end{tabular}} & UHD~\cite{xiao2022single} & 21.78 & 0.6198 \\
        \multicolumn{1}{c}{} 
                & MSDesmoking~\cite{wang2019multiscale} & 22.42 & 0.6415 \\
        \multicolumn{1}{c}{}
                & Wang \etal~\cite{wang2023surgical} & 23.63 & 0.6496 \\  
        \multicolumn{1}{c}{} 
                & MSBDN~\cite{dong2020multi} & 23.69 & 0.6515 \\ 
        \multicolumn{1}{c}{}
                & AODNet~\cite{li2017aod} & 23.45 & 0.6555 \\
        \multicolumn{1}{c}{} 
                & DADFNet~\cite{guo2023dadfnet} & 23.73 & 0.6516 \\
        \multicolumn{1}{c}{} 
                & DeHamar~\cite{guo2022image} & \pmb{24.12} & \pmb{0.6595} \\
        \midrule
        \multirow{7}{*}{\begin{tabular}[c]{@{}c@{}} Self-Supervised \\ Video Processing  \end{tabular}} 
                & BasicVSR~\cite{chan2021basicvsr} & 24.06 & 0.6560  \\
        \multicolumn{1}{c}{} 
                & BasicVSR++~\cite{chan2022basicvsr++} & 24.38 & \pmb{0.6592}  \\
        \multicolumn{1}{c}{} 
                & MAPNet~\cite{xu2023video} & 24.25 & 0.6567  \\
        \multicolumn{1}{c}{} & (Ours) SelfSVD-S & 24.84 & 0.6551\\
                & (Ours) SelfSVD$^\ast$-S & 24.25  & 0.6556 \\
                & (Ours) SelfSVD & \pmb{25.08} & 0.6564  \\
        \multicolumn{1}{c}{} 
                & (Ours) SelfSVD$^\ast$ & 24.62 & 0.6548 \\

		\bottomrule
	\end{tabular}
    \end{center}
\end{table*}

\section{Practical Deployment in Surgery}
\label{sec:supp_1}
In the main text, we have introduced how SelfSVD processes a single smoky video clip.
As a practical surgery video contains multiple smoky clips, here we illustrate how to handle it.
When processing the $i$-th smoky frame $\mathbf{S}_{i}$, an additional reference (Ref, $\mathbf{S}_{ref}$) should be fed into the model to help smoke removal.
There are two possible ways to select it.
One is to always adopt the starting frame, and another one is to select it dynamically as the surgery proceeds. 
The latter is more reasonable in surgical scenarios, as the changing video contents lead to starting frames providing insufficient information for long-distance ones.
Specifically, we prefer to select $\mathbf{S}_{ref}$ from previous neighboring frames that are clearer than $\mathbf{S}_{i}$.
As shown in \cref{alg:deployment}, we first deploy a Ref detector according to the residual between the current smoky input and the desmoking output.
$\mathbf{S}_{ref}$ is updated when next clearer frame occurs.
Then we feed $\mathbf{S}_{ref}$ and the smoky video clip $\{\mathbf{S}_{k}\}_{k=i}^{i+L}$ into a SelfSVD model, generating the clean results $\{\mathbf{\hat{I}}_{k}\}_{k=i}^{i+L}$.
$L$ is the frame number of the current video clip and we set it to 5.
Please see some visualization examples at the \href{https://github.com/ZcsrenlongZ/SelfSVD}{https://github.com/ZcsrenlongZ/SelfSVD}. 

\begin{algorithm*}[t!]
\caption{Pseudo code about the practical deployment of SelfSVD in surgery}
\begin{algorithmic}[1]
    \Require $\{\mathbf{S}_i\}_{i=1}^N$: $N$ surgery video frames,
    \For {{$i$ from $1$ to $N$ with stride $L$}}
            \If {$||\mathbf{S}_{i} - SelfSVD(\mathbf{S}_{i}, \mathbf{S}_{i})||_{1} < \mathbf{\epsilon}$}
            \Comment{utilize $\mathbf{S}_{i}$ as Ref for itself}
            \State 
                $\mathbf{S}_{ref}$ = $\mathbf{S}_{i}$ 
            \Comment{detect the additional reference input $\mathbf{S}_{ref}$}  
            \EndIf
            \State
            $\{\mathbf{\hat{I}}_{k}\}_{k=i}^{i+L}$ = $SelfSVD(\{\mathbf{S}_{k}\}_{k=i}^{i+L}, \mathbf{S}_{ref})$  
            \Comment{remove smoke for $\{\mathbf{S}_{k}\}_{k=i}^{i+L}$}
            \State
            \Return $\{\mathbf{\hat{I}}_{k}\}_{k=i}^{i+L}$.
    \EndFor
\end{algorithmic}
\label{alg:deployment}
\end{algorithm*}

\section{Limitation and Social Impact}
\label{sec:supp_7}
This work is still limited in processing complex surgery smoke droplets.
The droplets influence the accuracy of the alignment module, making it hard
to effectively utilize the complementary information from input frames.
It leads to some droplet traces in the desmoking results.

As for the social impact, this work is promising to be applied to laparoscopic surgery for observing the surgical sites more clearly.
It has no foreseeable negative influence.
Besides, the images utilized in this work are from professional hospitals and have been authorized to be public.
There is no personally identifiable information about patients or offensive content in the experimental data.

\clearpage

\clearpage


%
%
\bibliographystyle{splncs04}
\bibliography{egbib}

\end{document}